\documentclass[runningheads]{llncs}
\usepackage{graphicx}
\usepackage{comment}
\usepackage{amsmath,amssymb} %
\usepackage{color}

\usepackage[utf8]{inputenc}
\usepackage{makecell}
\usepackage{algorithm}
\usepackage[noend]{algpseudocode}
\usepackage{wrapfig,booktabs}
\usepackage[numbers,sort&compress]{natbib}
\usepackage{color}
\usepackage[normalem]{ulem}
\usepackage[table]{xcolor}
\usepackage{multirow}
\usepackage{mathtools}
\usepackage{dsfont}
\usepackage{comment}
\usepackage{enumitem,kantlipsum}
\usepackage{booktabs}
\usepackage{caption}
\usepackage{subcaption}
\usepackage{xspace}

\makeatletter
\@namedef{ver@everyshi.sty}{}
\makeatother

\newcommand{\red}[1]{\textcolor{red}{#1}}

\newcommand{\green}[1]{\textcolor{green}{#1}}

\DeclareUnicodeCharacter{2212}{-}
\begin{document}
\pagestyle{headings}
\mainmatter
\def\ECCVSubNumber{2813}

\title{Attention Guided Anomaly Localization in Images}

\titlerunning{Attention Guided Anomaly Localization in Images}

\author{Shashanka Venkataramanan$^{\star}$ \orcidID{0000-0003-1096-1342}, Kuan-Chuan Peng$^{\dagger}$ \orcidID{0000−0002−2682−9912}, Rajat Vikram Singh$^{\ddagger}$ \orcidID{0000-0002-1416-8344}, and Abhijit Mahalanobis$^{\star}$ \orcidID{0000-0002-2782-8655}}
\institute{$^{\star}$Center for Research in Computer Vision, University of Central Florida, Orlando, FL
$^{\dagger}$Mitsubishi Electric Research Laboratories, Cambridge, MA\\
$^{\ddagger}$Siemens Corporate Technology, Princeton, NJ\\
\email{ \tt\small shashankv@Knights.ucf.edu, kpeng@merl.com, singh.rajat@siemens.com, amahalan@crcv.ucf.edu}}

\authorrunning{S. Venkataramanan et al.}
\maketitle
\begin{abstract}
Anomaly localization is an important problem in computer vision which involves localizing anomalous regions within images with applications in industrial inspection, surveillance, and medical imaging. This task is challenging due to the small sample size and pixel coverage of the anomaly in real-world scenarios. Most prior works need to use anomalous training images to compute a class-specific threshold to localize anomalies. Without the need of anomalous training images, we propose Convolutional Adversarial Variational autoencoder with Guided Attention (CAVGA), which localizes the anomaly with a \textit{convolutional latent variable} to preserve the spatial information. In the unsupervised setting, we propose an \textit{attention expansion loss} where we encourage CAVGA to focus on all normal regions in the image.  Furthermore, in the weakly-supervised setting we propose a \textit{complementary guided attention loss}, where we encourage the attention map to focus on all normal regions while minimizing the attention map corresponding to anomalous regions in the image.  CAVGA outperforms the state-of-the-art (SOTA) anomaly localization methods on MVTec Anomaly Detection (MVTAD), modified ShanghaiTech Campus (mSTC) and Large-scale Attention based Glaucoma (LAG) datasets in the unsupervised setting and when using only 2\% anomalous images in the weakly-supervised setting. CAVGA also outperforms SOTA anomaly detection methods on the MNIST, CIFAR-10, Fashion-MNIST, MVTAD, mSTC and LAG datasets. 
\end{abstract}

\section{Introduction}
\label{introduction}

\begin{figure}[!ht]
  
    \begin{subfigure}[b]{0.6\linewidth}
    \includegraphics[width=1\linewidth]{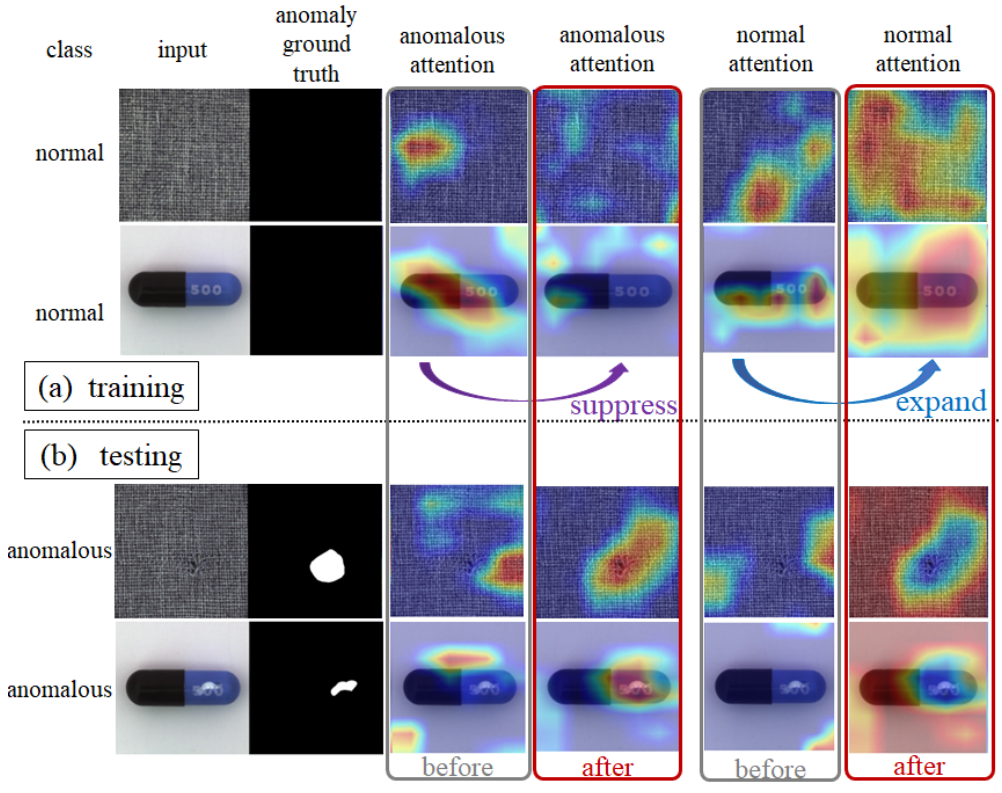}
    \caption{CAVGA main idea}
    \label{teaser}
    \end{subfigure}
    \hfil
    \begin{subfigure}[b]{0.35\linewidth}
    \scriptsize
    \centering
    \begin{tabular}{@{}c@{}c@{}c@{}}
    \toprule
    dataset &task &improvement\\
    \midrule
    MVTAD \cite{bergmann2019mvtec} &$l$ &(9/15; 4$\sim$85\%) \\ 
    MVTAD \cite{bergmann2019mvtec} &$d$ &(9/15; 2$\sim$30\%)\\
    mSTC \cite{liu2018future} &$l$ &(7/12; 2$\sim$42\%)\\
    mSTC \cite{liu2018future} &$d$ &(8/12; 1$\sim$38\%)\\
    LAG \cite{Li_2019_CVPR} &$l$ & (1/1; 16\%)\\
    LAG \cite{Li_2019_CVPR} &$d$ &(1/1; 1.1\%)\\
    MNIST \cite{lecun1998gradient} &$d$ &(8/10; 0.1$\sim$2.5\%)\\
    CIFAR-10 \cite{krizhevsky2009learning} &$d$ &(7/10; 3$\sim$31\%)\\
    F-MNIST \cite{xiao2017fashion} &$d$ &(8/10; 2$\sim$24\%)\\
    
    \bottomrule
    \\
    \multicolumn{3}{c}{$l$: localization; $d$: detection}\\
    \multicolumn{3}{c}{F-MNIST: Fashion-MNIST \cite{xiao2017fashion}}\\
    \\
    \multicolumn{3}{l}{$\bullet$ metric for $l$ : IoU}\\
    \multicolumn{3}{l}{$\bullet$ metric for $d$ in MVTAD, mSTC,}\\
    \multicolumn{3}{l}{and LAG: classification accuracy}\\
    \multicolumn{3}{l}{$\bullet$ metric for $d$ in MNIST,}\\
    \multicolumn{3}{l}{CIFAR-10 and Fashion-MNIST:}\\
    \multicolumn{3}{l}{area under ROC curve}\\
    \end{tabular} 
    \caption{improvement summary}
    \label{improvement_table}
    \end{subfigure}
    \caption{\subref{teaser} CAVGA uses the proposed complementary guided attention loss to encourage the attention map to cover the entire normal regions while suppressing the attention map corresponding to anomalous class in the training image. This enables the trained network to generate the anomalous attention map to localize the anomaly better at testing \subref{improvement_table} CAVGA's improvement over SOTA in the form of (number of outperforming/total categories; improvement (\%) in its metric)}
    
    \label{intro_fig}

\end{figure}

Recognizing whether an image is homogeneous with its previously observed distribution or whether it belongs to a novel or anomalous distribution has been identified as an important problem \cite{bergmann2019mvtec}. In this work, we focus on a related task, anomaly localization in images, which involves segmenting the anomalous regions within them. Anomaly localization has been applied in industrial inspection settings to segment defective product parts \cite{bergmann2019mvtec}, in surveillance to locate intruders \cite{nguyen2018weakly}, in medical imaging to segment tumor in brain MRI or glaucoma in retina images \cite{baur2018deep, Li_2019_CVPR}, etc. There has been an increase in analysis towards segmenting potential anomalous regions in images as acknowledged in \cite{dehaene2020iterative}.

Existing state-of-the-art (SOTA) anomaly localization methods \cite{sabokrou2018avid, bergmann2018improving} are based on deep learning. However, developing deep learning based algorithms for this task can be challenging due to the small pixel coverage of the anomaly and lack of suitable data since images with anomalies are rarely available in real-world scenarios \cite{bergmann2019mvtec}. Existing SOTA methods tackle this challenge using autoencoders \cite{sabokrou2018avid, dimokranitou2017adversarial} and GAN based approaches \cite{ravanbakhsh2019training, zenati2018efficient, akcay2018ganomaly}, which use a thresholded pixel-wise difference between the input and reconstructed image to localize anomalies. But, their methods need to determine class-specific thresholds using anomalous training images which can be unavailable in real-world scenarios.

To tackle these drawbacks of using anomalous training images, we propose Convolutional Adversarial Variational autoencoder with Guided Attention\\ (CAVGA), an unsupervised anomaly localization method which requires no anomalous training images. CAVGA comprises of a \textit{convolutional latent variable} to preserve the spatial relation between the input and latent variable. Since real-world applications may have access to only limited training data \cite{bergmann2019mvtec}, we propose to localize the anomalies by  using supervision on attention maps. This is motivated by the finding in \cite{li2018tell} that attention based supervision can alleviate the need of using large amount of training data. Intuitively, without any prior knowledge of the anomaly, humans need to look at the entire image to identify the anomalous regions. Based on this idea, we propose an \textit{attention expansion loss} where we encourage the network to generate an attention map that focuses on all normal regions of the image. 

Since annotating segmentation training data can be laborious \cite{Kimura_2020_WACV}, in the case when the annotator provides few anomalous training images without ground truth segmented anomalous regions, we extend CAVGA to a weakly supervised setting. Here, we introduce a classifier in CAVGA and propose a \textit{complementary guided attention loss} computed only for the normal images correctly predicted by the classifier. Using this complementary guided attention loss, we expand the normal attention but suppress the anomalous attention on the normal image, where normal/anomalous attention represents the areas affecting the classifier's normal/anomalous prediction identified by existing network visualization methods (e.g. Grad-CAM \cite{selvaraju2017grad}). Fig. \ref{intro_fig} \subref{teaser} (a) illustrates our attention mechanism during training, and Fig. \ref{intro_fig} \subref{teaser} (b) demonstrates that the resulting normal attention and anomalous attention on the anomalous testing images are visually complementary, which is consistent with our intuition. Furthermore, Fig. \ref{intro_fig} \subref{improvement_table} summarizes CAVGA's ability to outperform SOTA methods in anomaly localization on industrial inspection (MVTAD) \cite{bergmann2019mvtec}, surveillance (mSTC) \cite{liu2018future} and medical imaging (LAG) \cite{Li_2019_CVPR} datasets. We also show CAVGA's ability to outperform SOTA methods in anomaly detection on common benchmarks.

To the best of our knowledge, we are the first in anomaly localization to propose an end-to-end trainable framework with attention guidance which explicitly enforces the network to learn representations from the entire normal image. As compared to the prior works, our proposed approach CAVGA needs no anomalous training images to determine a class-specific threshold to localize the anomaly. Our contributions are:
\begin{itemize}
    \item \textbf{An attention expansion loss (\boldmath{$L_{ae}$})}, where we encourage the network to focus on the entire normal images in the unsupervised setting.
    \item \textbf{A complementary guided attention loss (\boldmath{$L_{cga}$})}, which we use to minimize the anomalous attention and simultaneously expand the normal attention for the normal images correctly predicted by the classifier.  
    \item \textbf{New SOTA}: In anomaly localization, CAVGA outperforms SOTA methods on the MVTAD and mSTC datasets in IoU and mean Area under ROC curve (AuROC) and also outperforms SOTA anomaly localization methods on LAG dataset in IoU. We also show CAVGA's ability to outperform SOTA methods for anomaly detection on the MVTAD, mSTC, LAG, MNIST \cite{lecun1998gradient}, CIFAR-10 \cite{krizhevsky2009learning} and Fashion-MNIST \cite{xiao2017fashion} datasets in classification accuracy. 
\end{itemize}

\section{Related Works}

Often used interchangeably, the terms anomaly localization and anomaly segmentation involve pixel-accurate segmentation of anomalous regions within an image \cite{bergmann2019mvtec}. They have been applied to industrial inspection settings to segment defective product parts \cite{bergmann2019mvtec}, medical imaging  to segment glaucoma in retina images \cite{Li_2019_CVPR}, etc. Image based anomaly localization has not been fully studied as compared to anomaly detection, where methods such as \cite{akcay2018ganomaly, schlegl2017unsupervised, bergmann2018improving,ravanbakhsh2019training, baur2018deep} employ a thresholded pixel wise difference between the input and reconstructed image to segment the anomalous regions. \cite{sabokrou2018avid} proposes an inpainter-detector network for patch-based localization in images. \cite{dehaene2020iterative} proposes gradient descent on a regularized autoencoder  while Liu \textit{et al.} \cite{liu2019towards} (denoted as ADVAE) generate gradient based attention maps from the latent space of the trained model.
We compare CAVGA with the existing methods relevant to anomaly localization in the unsupervised setting in Table \ref{table_related_works}  and show that among the listed methods, only CAVGA shows all the listed properties.


		

\begin{table}[t]
\begin{center}
\setlength{\tabcolsep}{4.0pt}
\scriptsize
\caption{
Comparison between CAVGA and other anomaly localization methods in the unsupervised setting in terms of the working properties. Among all the listed methods, only CAVGA satisfies all the listed properties 
} 
\label{table_related_works}
\begin{tabular}{ccccccc}
\toprule
		Does the method satisfy each property? &\cite{akcay2018ganomaly,schlegl2017unsupervised} & \cite{baur2018deep}  &\cite{sabokrou2018avid}  &\cite{wang2019pathology}  &\cite{dehaene2020iterative,liu2019towards}   &CAVGA\\
		& \cite{ravanbakhsh2019training, bergmann2018improving} & &  &\cite{smilkov2017smoothgrad}  &\cite{abati2019latent}  &\\
\midrule
		\textbf{not} using anomalous training images &\red{\textbf{N}} & \red{\textbf{N}} &\green{\textbf{Y}}  &\green{\textbf{\textbf{Y}}} &\green{\textbf{Y}} &\green{\textbf{Y}} \\ 
		
	\noindent localize \textbf{multiple} modes of anomalies &\green{\textbf{Y}} & \red{\textbf{N}} &\red{\textbf{N}}  &\red{\textbf{N}}  &\green{\textbf{Y}}  &\green{\textbf{Y}}\\ 
	\noindent \textbf{pixel} (not patch) based localization    &\green{\textbf{Y}} & \green{\textbf{Y}} &\red{\textbf{N}}  &\green{\textbf{Y}}  &\green{\textbf{Y}}   &\green{\textbf{Y}}\\ 
	\noindent use \textbf{convolutional latent variable} &\red{\textbf{N}} & \green{\textbf{Y}}  &\red{\textbf{N}}  &\red{\textbf{N}}  &\red{\textbf{N}}  &\green{\textbf{Y}}\\
\bottomrule
\end{tabular}
\end{center}
\end{table}

Anomaly detection involves determining an image as normal or anomalous \cite{akcay2018ganomaly}. One-class classification and anomaly detection are related to novelty detection \cite{perera2019ocgan} which has been widely studied in computer vision \cite{masana2018metric, hendrycks2018deep,akcay2018ganomaly, vu2019anomaly,napoletano2018anomaly} and applied to video analysis \cite{cheng2013abnormal}, remote sensing \cite{matteoli2014overview}, etc. With the advance in GANs \cite{goodfellow2014generative}, SOTA methods perform anomaly detection by generating realistic normal images during training \cite{Kimura_2020_WACV,radford2015unsupervised, sabokrou2018adversarially,schlegl2017unsupervised, higgins2017beta}.   \cite{deecke2018image} proposes to search the latent space of the generator for detecting anomalies. \cite{perera2019ocgan} introduces latent-space-sampling-based network with information-negative mining while \cite{li2019exploring} proposes normality score function based on capsule network’s activation and reconstruction error. \cite{abati2019latent} proposes a deep autoencoder that learns the distribution of latent representation through autoregressive procedure. Unlike  \cite{ruff2019deep, bian2019novel, wang2019inductive, daniel2019deep} where anomalous training images are used for anomaly detection, CAVGA does not need anomalous training images.

\section{Proposed Approach: CAVGA}
\label{approach}

\begin{figure}[t!]
\centering
\includegraphics[width= .92 \linewidth]{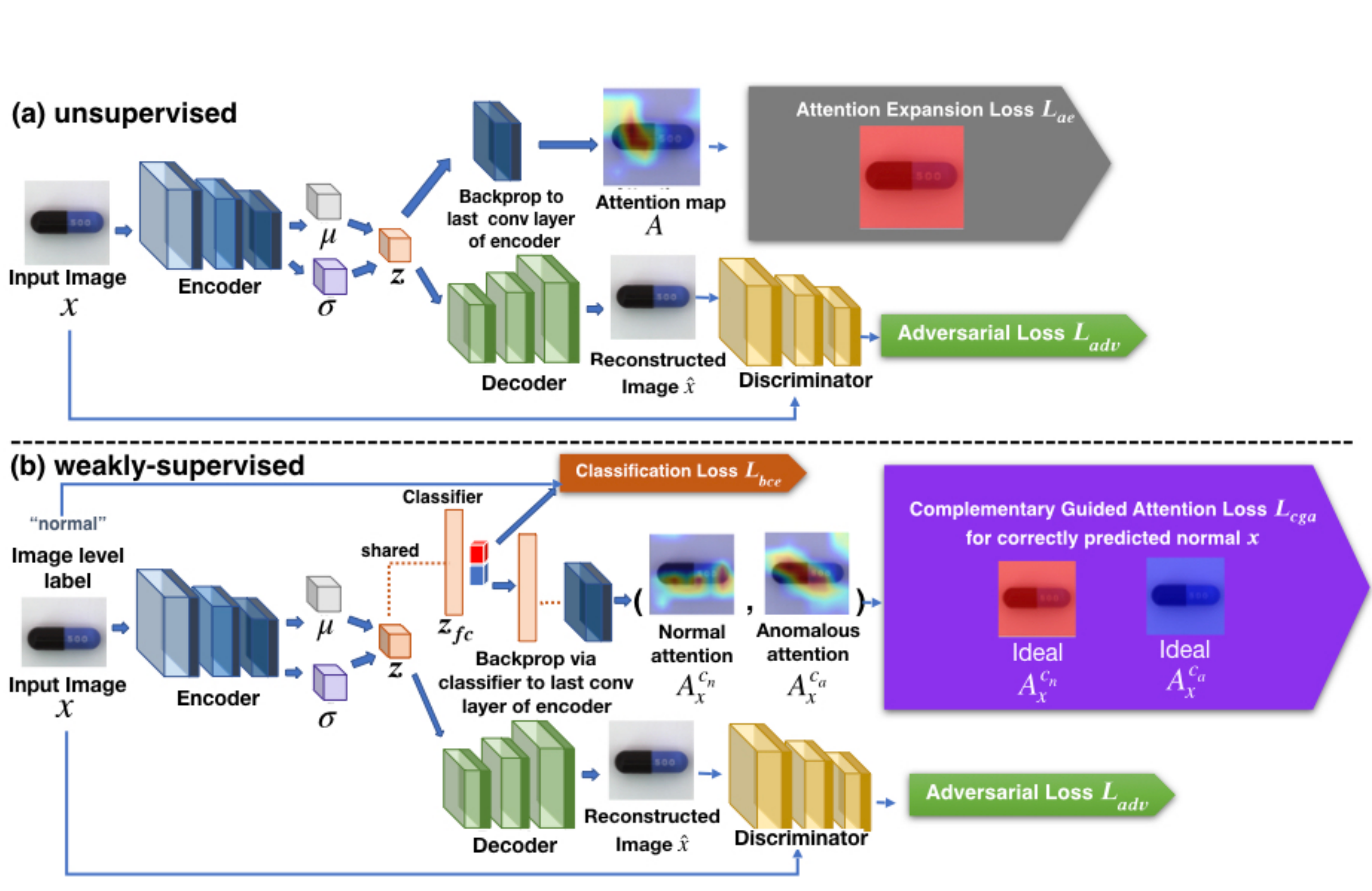}

\caption{
(a) The framework of CAVGA$_{u}$ where the attention expansion loss $L_{ae}$ guides the attention map $A$ computed from the latent variable $z$ to cover the entire normal image. (b) Illustration of CAVGA$_{w}$ with the complementary guided attention loss $L_{cga}$ to minimize the anomalous attention $A_{x}^{c_{a}}$ and expand the normal attention $A_{x}^{c_{n}}$ for the normal images correctly predicted by the classifier} 
\label{framework}

\end{figure}

\subsection{Unsupervised Approach: CAVGA$_u$}
\label{normal}
Fig. \ref{framework} (a) illustrates CAVGA in the unsupervised setting (denoted as CAVGA$_{u}$). CAVGA$_{u}$ comprises of a convolutional latent variable to preserve the spatial information between the input and latent variable. Since attention maps obtained from feature maps illustrate the regions of the image responsible for specific activation of neurons in the feature maps \cite{zagoruyko2016paying}, we propose an attention expansion loss such that the feature representation of the latent variable encodes all the normal regions. This loss encourages the attention map generated from the latent variable to cover the entire normal training image as illustrated in Fig. \ref{intro_fig} \subref{teaser} (a). During testing, we localize the anomaly from the areas of the image that the attention map does not focus on.

\textbf{Convolutional latent variable}
\label{avae}
Variational Autoencoder (VAE) \cite{kingma2013auto} is a generative model widely used for anomaly detection \cite{pawlowski2018unsupervised, kiran2018overview}. The loss function of training a vanilla VAE can be formulated as:
\begin{equation}
\label{eqn_recons}
L = L_{R}(x,\hat{x}) + KL(q_{\phi}(z|x)||p_{\theta}(z|x)),
\end{equation}
where $L_{R}(x,\hat{x}) = \frac{-1}{N}\sum_{i=1}^{N}x_{i}log(\hat{x}_{i}) + (1-x_{i})log(1-\hat{x}_{i})$ is the reconstruction loss between the input ($x$) and reconstructed images ($\hat{x}$), and $N$ is the total number of images. The posterior $p_{\theta}(z|x)$ is modeled using a standard Gaussian distribution prior $p(z)$ with the help of Kullback-Liebler ($KL$) divergence through $q_{\phi}(z|x)$. Since the vanilla VAE results in blurry reconstruction \cite{larsen2015autoencoding}, we use a discriminator ($D(.)$) to improve the stability of the training and generate sharper reconstructed images $\hat{x}$ using adversarial learning \cite{makhzani2015adversarial} formulated as follows:
\begin{equation}
\label{eqn_adv}
    L_{adv} =  - \frac{1}{N}\sum _{i=1} ^N \ log(D(x_{i})) + \ log(1 - D(\hat{x_{i}}))
\end{equation}

Unlike traditional autoencoders \cite{gutoski2017detection,bergmann2018improving} where the latent variable is flattened, inspired from \cite{baur2018deep}, we use a convolutional latent variable to preserve the spatial relation between the input and the latent variable. 

\textbf{Attention expansion loss $L_{ae}$}
\label{unsuper_guided_attn}
The main contribution of our work involves using supervision on attention maps to spatially localize the anomaly in the image. Most methods  \cite{schlegl2017unsupervised,vu2019anomaly,akcay2018ganomaly} employ a thresholded pixel-wise difference between the reconstructed image and the input image to localize the anomaly where the threshold is determined by using anomalous training images. However, CAVGA$_{u}$ learns to localize the anomaly using an attention map reflected through an end-to-end training process without the need of any anomalous training images. We use the feature representation of the latent variable $z$ to compute the attention map ($A$). $A$ is computed using Grad-CAM \cite{selvaraju2017grad} such that $A_{i,j} \in [0,1]$, where $A_{i,j}$ is the $(i, j)$ element of $A$.

Intuitively, $A$ obtained from feature maps focuses on the regions of the image based on the activation of neurons in the feature maps and its respective importance \cite{zhou2016learning,zagoruyko2016paying}. Due to the lack of prior knowledge about the anomaly, in general, humans need to look at the entire image to identify anomalous regions. We use this notion to learn the feature representation of the entire normal image by proposing an attention expansion loss, where we encourage the network to generate an attention map covering all the normal regions. This attention expansion loss for each image $L_{ae,1}$ is defined as:
\begin{equation}
\label{eqn_inv_attn}
    L_{ae,1} = \frac{1}{\left | A \right |}\sum_{i,j} (1 - A_{i,j})
\end{equation}

where $|A|$ is the total number of elements in $A$. The final attention expansion loss $L_{ae}$ is the average of $L_{ae,1}$ over the $N$ images. Since the idea of attention mechanisms involves locating the most salient regions in the image \cite{Li_2019_CVPR} which typically does not cover the entire image, we use $L_{ae}$ as an additional supervision on the network, such that the trained network generates an attention map that covers all the normal regions. Fig. \ref{intro_fig} \subref{teaser} (a) shows that before using $L_{ae}$ i.e. training CAVGA$_u$ only with adversarial learning ($L_{adv}$ + $L$) does not encode all the normal regions into the latent variable, and that the attention map fails to cover the entire image, which is overcome after using $L_{ae}$. Furthermore, supervising on attention maps prevents the trained model to make inference based on incorrect areas and also alleviates the need of using large amount of training data as shown in \cite{li2018tell}, which is not explicitly enforced in existing methods \cite{sabokrou2018avid, akcay2018ganomaly, bergmann2018improving}.

We form the final objective function $L_{final}$ below:
\begin{equation}
\label{eqn_finalunsuper}
    L_{final} = w_rL + w_{adv}L_{adv} + w_{ae}L_{ae},
\end{equation}
where $w_r$, $w_{adv}$, and $w_{ae}$ are empirically set as $1, 1$, and $0.01$ respectively.

During testing, we feed an image $x_{test}$ into the encoder followed by the decoder, which reconstructs an image $ \hat{x}_{test}$. As defined in \cite{schlegl2017unsupervised}, we compute the pixel-wise difference between $ \hat{x}_{test}$ and $x_{test}$ as the anomalous score $s_a$. Intuitively, if $x_{test}$ is drawn from the learnt distribution of $z$, then $s_{a}$ is small. Without using any anomalous training images in the unsupervised setting, we normalize $s_{a}$ between $[0,1]$ and empirically set 0.5 as the threshold to detect an image as anomalous. The attention map $A_{test}$ is computed from $z$ using Grad-CAM and is inverted (\textbf{1} - $A_{test}$) to obtain an anomalous attention map which localizes the anomaly. Here, \textbf{1} refers to a matrix of all ones with the same dimensions as $A_{test}$. We empirically choose 0.5 as the threshold on the anomalous attention map to evaluate the localization performance. 

\subsection{Weakly Supervised Approach: CAVGA$_w$}
\label{weakSuper}
CAVGA$_u$ can be further extended to a weakly supervised setting (denoted as CAVGA$_w$) where we explore the possibility of using few anomalous training images to improve the performance of anomaly localization. Given the labels of the anomalous and normal images without the pixel-wise annotation of the anomaly during training, we modify CAVGA$_u$ by introducing a binary classifier $C$ at the output of $z$ as shown in Fig. \ref{framework} (b) and train $C$ using the binary cross entropy loss $L_{bce}$. Given an image $x$ and its ground truth label $y$, we define $p \in \{c_{a}, c_{n}\}$ as the prediction of $C$, where $c_a$ and $c_n$ are anomalous and normal classes respectively. From Fig. \ref{framework} (b) we clone $z$ into a new tensor, flatten it to form a fully connected layer $z_{fc}$, and add a 2-node output layer to form $C$. $z$ and $z_{fc}$ share parameters.  Flattening $z_{fc}$ enables a higher magnitude of gradient backpropagation from $p$ \cite{selvaraju2017grad}. 

\textbf{Complementary guided attention loss $L_{cga}$} Although, attention maps generated from a trained classifier have been used in weakly supervised semantic segmentation tasks \cite{selvaraju2017grad,oquab2015object}, to the best of our knowledge, we are the first to propose supervision on attention maps for anomaly localization in the weakly supervised setting. Since the attention map depends on the performance of $C$ \cite{li2018tell}, we propose the complementary guided attention loss $L_{cga}$ based on $C$'s prediction to improve anomaly localization. We use Grad-CAM to compute the attention map for the anomalous class $A_x^{c_{a}}$ and the attention map for the normal class $A_x^{c_{n}}$ on the normal image $x$ ($y=c_n$). Using $A_x^{c_{a}}$ and $A_x^{c_{n}}$, we propose $L_{cga}$ where we minimize the areas covered by $A_x^{c_{a}}$ but simultaneously enforce $A_x^{c_{n}}$ to cover the entire normal image. Since the attention map is computed by backpropagating the gradients from $p$, any incorrect $p$ would generate an undesired attention map. This would lead to the network learning to focus on erroneous areas of the image during training, which we avoid using $L_{cga}$. We compute $L_{cga}$ only for the normal images correctly classified by the classifier i.e. if $p = y = c_n$. We define $L_{cga,1}$, the complementary guided attention loss for each image, in the weakly supervised setting as:
\begin{equation}
\label{eqn_abnormal}
    L_{cga,1} = \frac{\mathds{1}\left (p = y = c_n \right )}{\left | A_x^{c_n} \right |}\sum_{i,j}  (1 - (A_x^{c_n})_{i,j} + (A_x^{c_a})_{i,j}),
\end{equation}
where $\mathds{1}\left ( \cdot \right )$ is an indicator function. $L_{cga}$ is the average of $L_{cga,1}$ over the $N$ images. Our final objective function $L_{final}$ is defined as:
\begin{equation}
\label{eqn_weaksuper}
L_{final} = w_rL + w_{adv}L_{adv} + w_cL_{bce} + w_{cga}L_{cga},
\end{equation}
where $w_r$, $w_{adv}, w_{c}$, and $w_{cga}$ are empirically set as $1, 1, 0.001$, and $0.01$ respectively. During testing, we use $C$ to predict the input image $x_{test}$ as anomalous or normal. The anomalous attention map $A_{test}$ of $x_{test}$ is computed when $y = c_a$. We use the same evaluation method as that in Sec. \ref{unsuper_guided_attn} for anomaly localization.

\section{Experimental Setup}
\label{experiments}
\begin{table}[t]
\begin{center}
\renewcommand{\arraystretch}{0.9}
\setlength{\tabcolsep}{3.0pt}
\scriptsize
\caption{
Our experimental settings. Notations: $u$: unsupervised; $w$: weakly supervised; $D_M$: MNIST \cite{lecun1998gradient}; $D_F$: Fashion-MNIST \cite{xiao2017fashion}; $D_C$: CIFAR-10 \cite{krizhevsky2009learning}
}
\label{summary_datasets}
\begin{tabular}{ccccccccc}
\toprule
		  property $\backslash$ dataset             &\multicolumn{2}{c}{MVTAD \cite{bergmann2019mvtec}}   &\multicolumn{2}{c}{mSTC \cite{liu2018future}} 
		  &LAG \cite{Li_2019_CVPR} &$D_M$ &$D_F$ &$D_C$\\   
\midrule
		   setting            &$u$ &$w$ &$u$ &$w$ &$u$ &$u$ &$u$ &$u$\\
		   \# total classes &15 &15 &13 &13 &1 &10 &10 &10\\
		   \# normal training images        &3629                &3629          &244875        &244875 &2632 &$\sim$6k &6k &5k\\
		   \# anomalous training images        &0                &35          &0        &1763 &0 &0  &0 &0\\
		   \# normal testing images        &467 &467 &21147 &21147 &800 & $\sim$1k &1k &1k\\
		   \# anomalous testing images        &1223 &1223 &86404 &86404 & 2392 &$\sim$9k &9k &9k\\
\bottomrule
\end{tabular}
\end{center}

\end{table}

\textbf{Benchmark datasets:} We evaluate CAVGA on the MVTAD \cite{bergmann2019mvtec}, mSTC \cite{liu2018future} and  LAG \cite{Li_2019_CVPR} datasets for anomaly localization, and the MVTAD, mSTC, LAG, MNIST \cite{lecun1998gradient}, CIFAR-10 \cite{krizhevsky2009learning} and Fashion-MNIST \cite{xiao2017fashion} datasets for anomaly detection. Since  STC dataset \cite{liu2018future} is designed for video instead of image anomaly detection, we extract every $5^{\text{th}}$ frame of the video from each scene for training and testing without using any temporal information. We term the modified STC dataset as mSTC and summarize the experimental settings in Table \ref{summary_datasets}.

\textbf{Baseline methods:} For anomaly localization, we compare CAVGA with AVID \cite{sabokrou2018avid}, AE$_\text{L2}$ \cite{bergmann2018improving}, AE$_\text{SSIM}$ \cite{bergmann2018improving}, AnoGAN \cite{schlegl2017unsupervised}, CNN feature dictionary (CNNFD) \cite{napoletano2018anomaly}, texture inspection (TI) \cite{bottger2016real}, $\gamma$-VAE grad \cite{dehaene2020iterative} (denoted as $\gamma$-VAE$_{\text{g}}$), LSA \cite{abati2019latent}, ADVAE \cite{liu2019towards} and variation model (VM) \cite{steger2001similarity} based approaches on the MVTAD and mSTC datasets. Since \cite{dehaene2020iterative} does not provide the code for their method, we adapt the code from \cite{iclr20_code} and report its best result using our experimental settings. We also compare CAVGA$_u$ with CAM \cite{zhou2016learning}, GBP \cite{springenberg2014striving}, SmoothGrad \cite{smilkov2017smoothgrad} and Patho-GAN \cite{wang2019pathology} on the LAG dataset. In addition, we compare CAVGA$_u$ with LSA \cite{abati2019latent}, OCGAN \cite{perera2019ocgan}, ULSLM \cite{wolf2020unsupervised}, CapsNet PP-based and CapsNet RE-based \cite{li2019exploring} (denoted as CapsNet$_\text{PP}$ and CapsNet$_\text{RE}$), AnoGAN \cite{schlegl2017unsupervised}, ADGAN \cite{deecke2018image}, and $\beta$-VAE \cite{higgins2017beta} on the MNIST, CIFAR-10 and Fashion-MNIST datasets for anomaly detection.

\textbf{Architecture details:} Based on the framework in Fig. \ref{framework} (a), we use the convolution layers of ResNet-18 \cite{he2016deep} as our encoder pretrained from ImageNet \cite{russakovsky2015imagenet} and finetuned on each category / scene individually. Inspired from \cite{brock2018large}, we propose to use the residual generator as our residual decoder by modifying it with a convolution layer interleaved between two upsampling layers. The skip connection added from the output of the upsampling layer to the output of the convolution layer, increases mutual information between observations and latent variable and also avoids latent variable collapse \cite{dieng2019avoiding}. We use the discriminator of DC-GAN \cite{radford2015unsupervised} pretrained on the Celeb-A dataset \cite{liu2015faceattributes} and finetuned on our data as our discriminator and term this network as CAVGA-R. For fair comparisons with the baseline approaches in terms of network architecture, we use the discriminator and generator of DC-GAN pretrained on the Celeb-A dataset as our encoder and decoder respectively. We keep the same discriminator as discussed previously and term this network as CAVGA-D. CAVGA-D$_u$ and CAVGA-R$_u$ are termed as CAVGA$_u$ in the unsupervised setting, and CAVGA-D$_w$ and CAVGA-R$_w$ as CAVGA$_w$ in weakly supervised setting respectively.

\textbf{Training and evaluation:} For anomaly localization and detection on the MVTAD, mSTC and LAG datasets, the network is trained only on normal images in the unsupervised setting. In the weakly supervised setting, since none of the baseline methods provide the number of anomalous training images they use to compute the threshold, we randomly choose 2\% of the anomalous images along with all the normal training images for training. On the MNIST, CIFAR-10 and Fashion-MNIST datasets, we follow the same procedure as defined in \cite{deecke2018image} (training/testing uses single class as normal and the rest of the classes as anomalous. We train CAVGA-D$_{u}$ using this normal class).
For anomaly localization, we show the AuROC \cite{bergmann2019mvtec} and the Intersection-over-Union (IoU) between the generated attention map and the ground truth. Following \cite{bergmann2019mvtec}, we use the mean of accuracy of correctly classified anomalous images and normal images to evaluate the performance of anomaly detection on both the normal and anomalous images on the MVTAD, mSTC and LAG datasets. On the MNIST, CIFAR-10, and Fashion-MNIST datasets, same as \cite{deecke2018image}, we use AuROC for evaluation.

\begin{table}[t]
\begin{center}
\renewcommand{\arraystretch}{.9}
\setlength{\tabcolsep}{1.0pt}
\caption{Performance comparison of anomaly localization in category-specific IoU, mean IoU ($\overline{\text{IoU}}$), and mean AuROC ($\overline{\text{AuROC}}$) on the MVTAD dataset. The darker cell color indicates better performance ranking in each row} 
\label{table_iou_mvtec}
\scriptsize
\begin{tabular}{cccccccccccc}
\toprule
		Category &AVID &AE$_\text{SSIM}$ &AE$_\text{L2}$ &AnoGAN &$\gamma$-VAE$_{\text{g}}$ &LSA &ADVAE  &CAVGA &CAVGA &CAVGA &CAVGA\\
		&\cite{sabokrou2018avid} &\cite{bergmann2018improving} &\cite{bergmann2018improving} &\cite{schlegl2017unsupervised} &\cite{dehaene2020iterative} &\cite{abati2019latent} &\cite{liu2019towards}    &-D$_{u}$ &-R$_{u}$ &-D$_{w}$ &-R$_{w}$\\
\midrule
		Bottle   &\cellcolor{blue!40}0.28 &\cellcolor{blue!20}0.15  &\cellcolor{blue!25}0.22 &\cellcolor{blue!10}0.05 &\cellcolor{blue!30}0.27 &\cellcolor{blue!30}0.27    &\cellcolor{blue!30}0.27  &\cellcolor{blue!50}0.30  &\cellcolor{blue!60}0.34   &\cellcolor{blue!70}0.36 &\cellcolor{blue!80}\color{white}0.39\\
		Hazelnut  &\cellcolor{blue!50}0.54 &\cellcolor{blue!10}0.00  &\cellcolor{blue!20}0.41 &\cellcolor{blue!15}0.02 &\cellcolor{blue!70}0.63 &\cellcolor{blue!20}0.41    &\cellcolor{blue!30}0.44  &\cellcolor{blue!30}0.44 &\cellcolor{blue!40}0.51 &\cellcolor{blue!60}0.58 &\cellcolor{blue!80}\color{white}0.79 \\
		Capsule   &\cellcolor{blue!25}0.21 &\cellcolor{blue!10}0.09  &\cellcolor{blue!20}0.11 &\cellcolor{blue!5}0.04 &\cellcolor{blue!40}0.24 &\cellcolor{blue!30}0.22   &\cellcolor{blue!20}0.11  &\cellcolor{blue!50}0.25  &\cellcolor{blue!60}0.31 &\cellcolor{blue!70}0.38 &\cellcolor{blue!80}\color{white}0.41\\
		Metal Nut &\cellcolor{blue!15}0.05 &\cellcolor{blue!10}0.01  &\cellcolor{blue!30}0.26 &\cellcolor{blue!5}0.00 &\cellcolor{blue!25}0.22 &\cellcolor{blue!40}0.38    &\cellcolor{blue!80}\color{white}0.49  &\cellcolor{blue!50}0.39  &\cellcolor{blue!60}0.45  &\cellcolor{blue!70}0.46 &\cellcolor{blue!70}0.46\\
		
		Leather  &\cellcolor{blue!15}0.32 &\cellcolor{blue!20}0.34  &\cellcolor{blue!30}0.67 &\cellcolor{blue!20}0.34 &\cellcolor{blue!25}0.41 &\cellcolor{blue!50}0.77 &\cellcolor{blue!5}0.24  &\cellcolor{blue!40}0.76 &\cellcolor{blue!60}0.79  &\cellcolor{blue!70}0.80 &\cellcolor{blue!80}\color{white}0.84\\
		Pill    &\cellcolor{blue!15}0.11  &\cellcolor{blue!10}0.07  &\cellcolor{blue!30}0.25 &\cellcolor{blue!20}0.17 &\cellcolor{blue!70}0.48 &\cellcolor{blue!25}0.18    &\cellcolor{blue!25}0.18  &\cellcolor{blue!40}0.34  &\cellcolor{blue!50}0.40 &\cellcolor{blue!60}0.44 &\cellcolor{blue!80}\color{white}0.53\\
		Wood    &\cellcolor{blue!10}0.14    &\cellcolor{blue!25}0.36  &\cellcolor{blue!20}0.29 &\cellcolor{blue!10}0.14 &\cellcolor{blue!40}0.45 &\cellcolor{blue!30}0.41 &\cellcolor{blue!10}0.14  &\cellcolor{blue!50}0.56  &\cellcolor{blue!60}0.59 &\cellcolor{blue!70}0.61 &\cellcolor{blue!80}\color{white}0.66\\
		
		Carpet &\cellcolor{blue!10}0.25   &\cellcolor{blue!25}0.69  &\cellcolor{blue!20}0.38 &\cellcolor{blue!15}0.34 &\cellcolor{blue!70}0.79 &\cellcolor{blue!60}0.76 &\cellcolor{blue!5}0.10  &\cellcolor{blue!40}0.71  &\cellcolor{blue!50}0.73 &\cellcolor{blue!30}0.70 &\cellcolor{blue!80}\color{white}0.81\\
		Tile  &\cellcolor{blue!20}0.09    &\cellcolor{blue!5}0.04  &\cellcolor{blue!30}0.23 &\cellcolor{blue!10}0.08 &\cellcolor{blue!60}0.38 &\cellcolor{blue!50}0.32 &\cellcolor{blue!30}0.23  &\cellcolor{blue!40}0.31  &\cellcolor{blue!60}0.38 &\cellcolor{blue!70}0.47 &\cellcolor{blue!80}\color{white}0.81\\
		Grid  &\cellcolor{blue!50}0.51    &\cellcolor{blue!80}\color{white}0.88  &\cellcolor{blue!70}0.83 &\cellcolor{blue!10}0.04 &\cellcolor{blue!25}0.36 &\cellcolor{blue!15}0.20 &\cellcolor{blue!5}0.02 &\cellcolor{blue!20}0.32  &\cellcolor{blue!30}0.38 &\cellcolor{blue!40}0.42 &\cellcolor{blue!60}0.55\\
		Cable &\cellcolor{blue!30}0.27    &\cellcolor{blue!10}0.01  &\cellcolor{blue!15}0.05 &\cellcolor{blue!10}0.01 &\cellcolor{blue!25}0.26 &\cellcolor{blue!40}0.36    &\cellcolor{blue!20}0.18  &\cellcolor{blue!50}0.37  &\cellcolor{blue!60}0.44 &\cellcolor{blue!70}0.49 &\cellcolor{blue!80}\color{white}0.51\\
		Transistor &\cellcolor{blue!20}0.18 &\cellcolor{blue!10}0.01  &\cellcolor{blue!30}0.22 &\cellcolor{blue!15}0.08 &\cellcolor{blue!70}0.44 &\cellcolor{blue!25}0.21    &\cellcolor{blue!40}0.30  &\cellcolor{blue!40}0.30  &\cellcolor{blue!50}0.35  &\cellcolor{blue!60}0.38 &\cellcolor{blue!80}\color{white}0.45\\
		Toothbrush &\cellcolor{blue!25}0.43 &\cellcolor{blue!10}0.08  &\cellcolor{blue!40}0.51 &\cellcolor{blue!5}0.07 &\cellcolor{blue!20}0.37 &\cellcolor{blue!30}0.48    &\cellcolor{blue!15}0.14  &\cellcolor{blue!50}0.54 &  \cellcolor{blue!60}0.57 &\cellcolor{blue!70}0.60 &\cellcolor{blue!80}\color{white}0.63\\
		
		Screw &\cellcolor{blue!25}0.22    &\cellcolor{blue!15}0.03  &\cellcolor{blue!30}0.34 &\cellcolor{blue!10}0.01 &\cellcolor{blue!40}0.38 &\cellcolor{blue!40}0.38    &\cellcolor{blue!20}0.17  &\cellcolor{blue!50}0.42 &  \cellcolor{blue!60}0.48 &\cellcolor{blue!70}0.51 &\cellcolor{blue!80}\color{white}0.66\\
		Zipper &\cellcolor{blue!50}0.25    &\cellcolor{blue!15}0.10  &\cellcolor{blue!20}0.13 &\cellcolor{blue!5}0.01 &\cellcolor{blue!30}0.17 &\cellcolor{blue!25}0.14    &\cellcolor{blue!10}0.06 &\cellcolor{blue!40}0.20  &\cellcolor{blue!60}0.26 &\cellcolor{blue!70}0.29 &\cellcolor{blue!80}\color{white}0.31\\
\midrule
$\overline{\text{IoU}}$ &\cellcolor{blue!20}0.26  &\cellcolor{blue!10}0.19 &\cellcolor{blue!25}0.33 &\cellcolor{blue!5}0.09 &\cellcolor{blue!40}0.39 &\cellcolor{blue!30}0.37 &\cellcolor{blue!15}0.20  &\cellcolor{blue!50}0.41 &\cellcolor{blue!60}0.47 &\cellcolor{blue!70}0.50 &\cellcolor{blue!80}\color{white}0.59\\
\midrule
$\overline{\text{AuROC}}$ &\cellcolor{blue!15}0.78 &\cellcolor{blue!50}0.87 &\cellcolor{blue!25}0.82 &\cellcolor{blue!10}0.74 &\cellcolor{blue!40}0.86 &\cellcolor{blue!20}0.79 &\cellcolor{blue!40}0.86  &\cellcolor{blue!30}0.85 &\cellcolor{blue!60}0.89 &\cellcolor{blue!70}0.92 &\cellcolor{blue!80}\color{white}0.93\\
\bottomrule
\end{tabular}
\end{center}
\end{table}

\section{Experimental Results}
\label{comparisions_details}

We use the cell color in the quantitative result tables to denote the performance ranking in that row, where darker cell color means better performance.

\begin{figure}[h!]
\centering

\includegraphics[width= .75\linewidth]{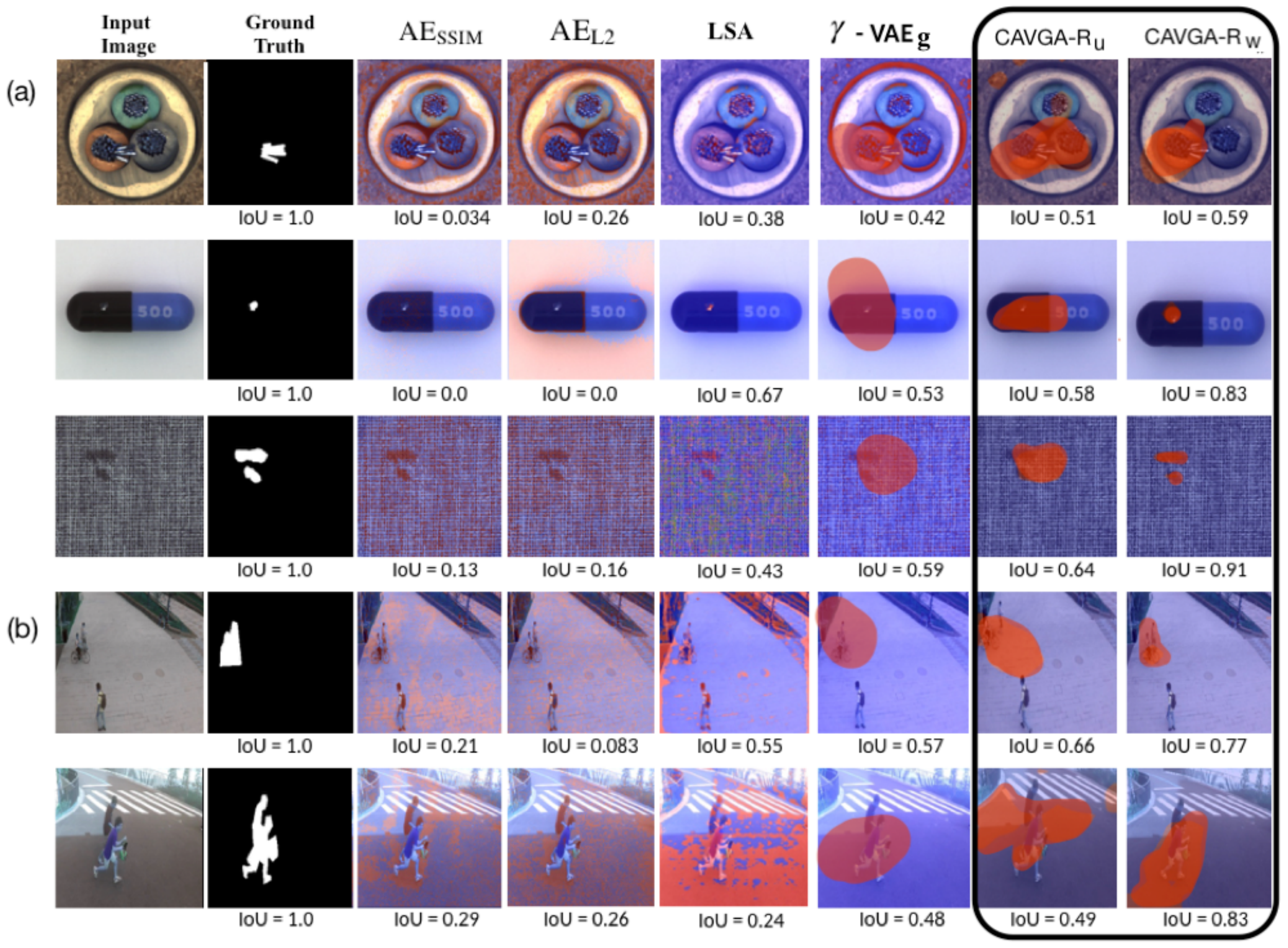}
\caption{
Qualitative results on (a) MVTAD \& (b) mSTC datasets respectively. The anomalous attention map (in red) depicts the localization of the anomaly}

\label{qual_mvtec}
\end{figure}

\begin{figure}[h!]
\centering
\includegraphics[width= .8\linewidth]{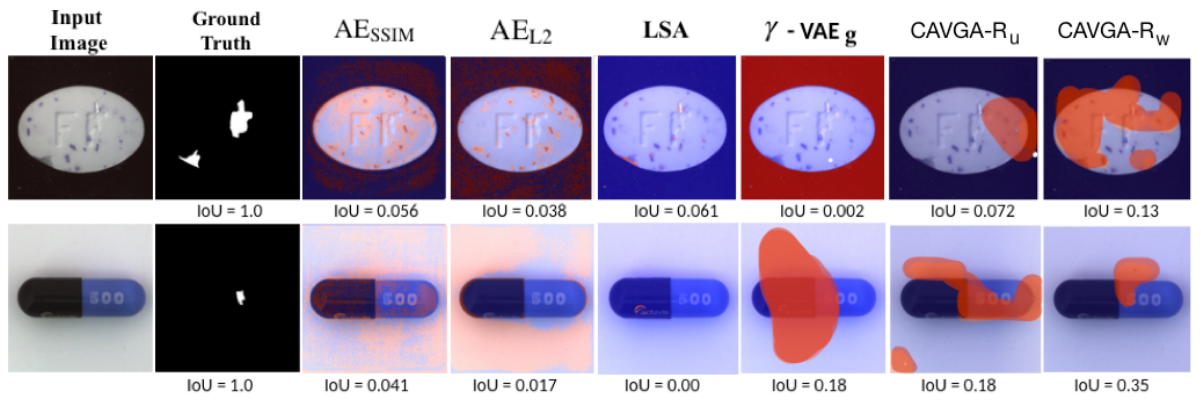}
\caption{
Examples of incorrect localization of the anomaly on the MVTAD dataset by CAVGA-R$_{u}$ and CAVGA-R$_{w}$
}
\label{qual_failure}

\end{figure}

\textbf{Performance on anomaly localization:} Fig. \ref{qual_mvtec} (a) shows the qualitative results and Table \ref{table_iou_mvtec} shows that CAVGA$_{u}$ localizes the anomaly better compared to the baselines on the MVTAD dataset. CAVGA-D$_{u}$ outperforms the best performing baseline method ($\gamma$-VAE$_{\text{g}}$) in mean IoU by 5\%. Most baselines use anomalous training images to compute class-specific threshold to localize anomalies. \textit{Needing no anomalous training images}, CAVGA-D$_{u}$ still outperforms all the mentioned baselines in mean IoU. In terms of mean AuROC, CAVGA-D$_{u}$ outperforms CNNFD, TI and VM by 9\%, 12\% and 10\% respectively and achieves comparable results with best baseline method.  Table \ref{table_iou_mvtec} also shows that CAVGA-D$_{w}$ outperforms CAVGA-D$_{u}$ by 22\% and 8\% on mean IoU and mean AuROC respectively. CAVGA-D$_{w}$ also outperforms the baselines in mean AuROC. Fig. \ref{qual_failure} illustrates that one challenge in anomaly localization is the  low contrast between the anomalous regions and their background. In such scenarios, although still outperforming the baselines, CAVGA does not localize the anomaly well.

\begin{table}[h!]
\begin{center}
\renewcommand{\arraystretch}{.8}
\setlength{\tabcolsep}{1.6pt}
\caption{Performance comparison of anomaly localization in IoU and its mean ($\overline{\text{IoU}}$) along with anomaly detection in terms of mean of accuracy of correctly classified anomalous images and normal images on the mSTC dataset for each scene ID $s_i$. For anomaly localization, we also list the mean AuROC ($\overline{\text{AuROC}}$)}
\label{table_iou_st}
\scriptsize
\begin{tabular}{ccccccccccc}
\toprule
		  Task $\backslash$ Method &$s_i$ & \scriptsize $\gamma$-VAE$_{\text{g}}$ &\scriptsize AVID &\scriptsize LSA &\scriptsize AE$_\text{SSIM}$ &\scriptsize AE$_\text{L2}$ &\scriptsize CAVGA &\scriptsize CAVGA &\scriptsize CAVGA      &\scriptsize CAVGA \\
		   & & \cite{dehaene2020iterative} &\cite{sabokrou2018avid} & \cite{abati2019latent} & \cite{bergmann2018improving} &\cite{bergmann2018improving} &-D$_{u}$ &-R$_{u}$  &-D$_{w}$      &-R$_{w}$ \\
		  
\midrule
		& 01 &\cellcolor{blue!30}0.239 &\cellcolor{blue!10}0.182 &\cellcolor{blue!40}0.244  &\cellcolor{blue!20}0.201  &\cellcolor{blue!5}0.163  &\cellcolor{blue!50}0.267 &\cellcolor{blue!60}0.316   &\cellcolor{blue!70}0.383 &\cellcolor{blue!80}\color{white}0.441 \\
		& 02 &\cellcolor{blue!50}0.206  &\cellcolor{blue!50}0.206  &\cellcolor{blue!30}0.183  &\cellcolor{blue!10}0.081  &\cellcolor{blue!20}0.172  &\cellcolor{blue!40}0.190  &\cellcolor{blue!60}0.234   &\cellcolor{blue!70}0.257 &\cellcolor{blue!80}\color{white}0.349 \\
		& 03 &\cellcolor{blue!40}0.272  &\cellcolor{blue!10}0.162  &\cellcolor{blue!30}0.265  &\cellcolor{blue!15}0.218  &\cellcolor{blue!25}0.240  &\cellcolor{blue!50}0.277  &\cellcolor{blue!60}0.293  &\cellcolor{blue!70}0.313 &\cellcolor{blue!80}\color{white}0.465 \\
		& 04  &\cellcolor{blue!50}0.290  &\cellcolor{blue!20}0.263  &\cellcolor{blue!30}0.271  &\cellcolor{blue!5}0.118  &\cellcolor{blue!10}0.125  &\cellcolor{blue!40}0.283  &\cellcolor{blue!60}0.349  &\cellcolor{blue!70}0.360 &\cellcolor{blue!80}\color{white}0.381 \\
		
		& 05 &\cellcolor{blue!60}0.318  &\cellcolor{blue!20}0.234  &\cellcolor{blue!30}0.287  &\cellcolor{blue!10}0.162  &\cellcolor{blue!5}0.129  &\cellcolor{blue!40}0.291  &\cellcolor{blue!50}0.312  &\cellcolor{blue!70}0.408 &\cellcolor{blue!80}\color{white}0.478 \\
		
		Localization & 06  &\cellcolor{blue!40}0.337  &\cellcolor{blue!30}0.314  &\cellcolor{blue!20}0.238  &\cellcolor{blue!10}0.215  &\cellcolor{blue!5}0.198  &\cellcolor{blue!50}0.344  &\cellcolor{blue!60}0.420  &\cellcolor{blue!70}0.455 &\cellcolor{blue!80}\color{white}0.589  \\
		
		& 07 &\cellcolor{blue!20}0.168  &\cellcolor{blue!50}0.214  &\cellcolor{blue!5}0.137  &\cellcolor{blue!30}0.191  &\cellcolor{blue!10}0.165  &\cellcolor{blue!40}0.198  &\cellcolor{blue!60}0.241  &\cellcolor{blue!70}0.284 &\cellcolor{blue!80}\color{white}0.366 \\
		
		& 08 &\cellcolor{blue!40}0.220  &\cellcolor{blue!20}0.168  &\cellcolor{blue!50}0.233  &\cellcolor{blue!10}0.069  &\cellcolor{blue!5}0.056  &\cellcolor{blue!30}0.219  &\cellcolor{blue!60}0.254  &\cellcolor{blue!70}0.295 &\cellcolor{blue!80}\color{white}0.371 \\
		
		& 09 &\cellcolor{blue!20}0.174 &\cellcolor{blue!40}0.193  &\cellcolor{blue!30}0.187  &\cellcolor{blue!10}0.038  &\cellcolor{blue!5}0.021  &\cellcolor{blue!50}0.247  &\cellcolor{blue!60}0.284  &\cellcolor{blue!70}0.313 &\cellcolor{blue!80}\color{white}0.365 \\
		& 10 &\cellcolor{blue!40}0.146  &\cellcolor{blue!20}0.137  &\cellcolor{blue!40}0.146  &\cellcolor{blue!10}0.116  &\cellcolor{blue!30}0.141  &\cellcolor{blue!50}0.149  &\cellcolor{blue!60}0.166  &\cellcolor{blue!70}0.245 &\cellcolor{blue!80}\color{white}0.295 \\
		
		& 11 &\cellcolor{blue!30}0.277 &\cellcolor{blue!20}0.264  &\cellcolor{blue!40}0.286  &\cellcolor{blue!10}0.101  &\cellcolor{blue!5}0.075  &\cellcolor{blue!50}0.309  &\cellcolor{blue!60}0.372  &\cellcolor{blue!70}0.441 &\cellcolor{blue!80}\color{white}0.588 \\
		
		& 12 &\cellcolor{blue!30}0.162  &\cellcolor{blue!50}0.180  &\cellcolor{blue!10}0.108  &\cellcolor{blue!60}0.203  &\cellcolor{blue!40}0.164  &\cellcolor{blue!5}0.098  &\cellcolor{blue!20}0.141  &\cellcolor{blue!70}0.207 &\cellcolor{blue!80}\color{white}0.263\\
\midrule
& $\overline{\text{IoU}}$ &\cellcolor{blue!40}0.234 &\cellcolor{blue!20}0.210  &\cellcolor{blue!30}0.215  &\cellcolor{blue!10}0.143 &\cellcolor{blue!5}0.137 &\cellcolor{blue!50}0.239 &\cellcolor{blue!60}0.281 &\cellcolor{blue!70}0.330 &\cellcolor{blue!80}\color{white}0.412\\
\midrule
& $\overline{\text{AuROC}}$ &\cellcolor{blue!40}0.82 &\cellcolor{blue!20}0.77 &\cellcolor{blue!30}0.81  &\cellcolor{blue!10}0.76 &\cellcolor{blue!5}0.74 &\cellcolor{blue!50}0.83 &\cellcolor{blue!60}0.85 &\cellcolor{blue!70}0.89 &\cellcolor{blue!80}\color{white}0.90\\
\midrule
        & 01 &\cellcolor{blue!40}0.75  &\cellcolor{blue!20}0.68  &\cellcolor{blue!40}0.75  &\cellcolor{blue!10}0.65  &\cellcolor{blue!30}0.72   &\cellcolor{blue!50}0.77   &\cellcolor{blue!70}0.85   &\cellcolor{blue!60}0.84 &\cellcolor{blue!80}\color{white}0.87 \\
        
		& 02 &\cellcolor{blue!20}0.75  &\cellcolor{blue!20}0.75  &\cellcolor{blue!50}0.79  &\cellcolor{blue!10}0.70  &\cellcolor{blue!5}0.61   &\cellcolor{blue!40}0.76  &\cellcolor{blue!60}0.84   &\cellcolor{blue!70}0.89 &\cellcolor{blue!80}\color{white}0.90 \\
		& 03  &\cellcolor{blue!40}0.81 &\cellcolor{blue!10}0.68  &\cellcolor{blue!5}0.63  &\cellcolor{blue!30}0.79  &\cellcolor{blue!20}0.71   &\cellcolor{blue!50}0.82   &\cellcolor{blue!60}0.84   &\cellcolor{blue!70}0.86&\cellcolor{blue!80}\color{white}0.88 \\
		
		& 04 &\cellcolor{blue!80}\color{white}0.83  &\cellcolor{blue!20}0.71  &\cellcolor{blue!30}0.79  &\cellcolor{blue!60}0.81  &\cellcolor{blue!10}0.66   &\cellcolor{blue!40}0.80   &\cellcolor{blue!40}0.80   &\cellcolor{blue!60}0.81 &\cellcolor{blue!80}\color{white}0.83 \\
		
		& 05 &\cellcolor{blue!60}0.86  &\cellcolor{blue!10}0.59  &\cellcolor{blue!30}0.68  &\cellcolor{blue!40}0.71  &\cellcolor{blue!20}0.67   &\cellcolor{blue!50}0.81   &\cellcolor{blue!60}0.86   &\cellcolor{blue!70}0.90 &\cellcolor{blue!80}\color{white}0.94 \\
		
		Detection& 06 &\cellcolor{blue!30}0.59  &\cellcolor{blue!40}0.62  &\cellcolor{blue!20}0.58  &\cellcolor{blue!5}0.47  &\cellcolor{blue!10}0.55   &\cellcolor{blue!50}0.64  &\cellcolor{blue!70}0.67   &\cellcolor{blue!60}0.65 &\cellcolor{blue!80}\color{white}0.70 \\
		
		& 07 &\cellcolor{blue!20}0.59 &\cellcolor{blue!50}0.63  &\cellcolor{blue!50}0.63  &\cellcolor{blue!10}0.36  &\cellcolor{blue!20}0.59   &\cellcolor{blue!30}0.60  &\cellcolor{blue!60}0.64   &\cellcolor{blue!70}0.75 &\cellcolor{blue!80}\color{white}0.77 \\
		
		& 08 &\cellcolor{blue!70}0.77  &\cellcolor{blue!30}0.73  &\cellcolor{blue!50}0.75  &\cellcolor{blue!10}0.69  &\cellcolor{blue!20}0.70   &\cellcolor{blue!40}0.74  &\cellcolor{blue!40}0.74   &\cellcolor{blue!60}0.76 &\cellcolor{blue!80}\color{white}0.80 \\
		
		& 09 &\cellcolor{blue!60}0.89 &\cellcolor{blue!50}0.88  &\cellcolor{blue!20}0.79  &\cellcolor{blue!30}0.84  &\cellcolor{blue!10}0.73   &\cellcolor{blue!40}0.87   &\cellcolor{blue!50}0.88   &\cellcolor{blue!70}0.90 &\cellcolor{blue!80}\color{white}0.91 \\
		
		& 10  &\cellcolor{blue!10}0.64  &\cellcolor{blue!20}0.80  &\cellcolor{blue!40}0.84  &\cellcolor{blue!30}0.83  &\cellcolor{blue!50}0.88   &\cellcolor{blue!50}0.88   &\cellcolor{blue!60}0.92   &\cellcolor{blue!80}\color{white}0.94 &\cellcolor{blue!80}\color{white}0.94 \\
		
		& 11 &\cellcolor{blue!40}0.78 &\cellcolor{blue!10}0.68  &\cellcolor{blue!20}0.71  &\cellcolor{blue!20}0.71  &\cellcolor{blue!30}0.75   &\cellcolor{blue!50}0.79   &\cellcolor{blue!60}0.81   &\cellcolor{blue!80}\color{white}0.83 &\cellcolor{blue!80}\color{white}0.83 \\
		
		& 12  &\cellcolor{blue!40}0.71  &\cellcolor{blue!30}0.66  &\cellcolor{blue!10}0.63  &\cellcolor{blue!20}0.65  &\cellcolor{blue!5}0.52   &\cellcolor{blue!50}0.76   &\cellcolor{blue!60}0.79   &\cellcolor{blue!70}0.81 &\cellcolor{blue!80}\color{white}0.83 \\
\midrule
& avg &\cellcolor{blue!40}0.75 &\cellcolor{blue!20}0.70  &\cellcolor{blue!30}0.71  &\cellcolor{blue!10}0.68 &\cellcolor{blue!5}0.67 &\cellcolor{blue!50}0.77 &\cellcolor{blue!60}0.80 &\cellcolor{blue!70}0.83 &\cellcolor{blue!80}\color{white}0.85\\
\bottomrule
\end{tabular}
\end{center}
\end{table}

\begin{table}[h!]
\begin{center}
\setlength{\tabcolsep}{2.0pt}
\scriptsize
\caption{Performance comparison of anomaly localization in IoU along with anomaly detection in terms of classification accuracy on the LAG dataset \cite{Li_2019_CVPR}
}
\label{table_lag}
\begin{tabular}{cccccc}
\toprule
		  Task $\backslash$ Method   & CAM \cite{zhou2016learning} & GBP \cite{springenberg2014striving}  &SmoothGrad \cite{smilkov2017smoothgrad} & Patho-GAN \cite{wang2019pathology} &CAVGA-D$_{u}$\\
		  
\midrule
		 Localization   &\cellcolor{blue!20}0.13 &\cellcolor{blue!5}0.09  &\cellcolor{blue!40}0.14 &\cellcolor{blue!60}0.37  &\cellcolor{blue!80} \color{white}0.43 \\ 
		 Detection      &\cellcolor{blue!5}0.68  &\cellcolor{blue!40}0.84   &\cellcolor{blue!20}0.79  &\cellcolor{blue!60}0.89  &\cellcolor{blue!80} \color{white}0.90 \\ 
\bottomrule
\end{tabular}
\end{center}
\end{table}

\begin{table}[h]
\begin{center}
\renewcommand{\arraystretch}{.9}
\setlength{\tabcolsep}{2.0pt}
\caption{The mean of accuracy of correctly classified anomalous images and normal images in anomaly detection on the MVTAD dataset }
\label{table_accu_mvtec}

\scriptsize
\begin{tabular}{ccccccccccc}
\toprule
		Category &AVID  &AE$_\text{SSIM}$  &AE$_\text{L2}$ &AnoGAN  &$\gamma$-VAE$_{\text{g}}$  &LSA  &CAVGA &CAVGA &CAVGA &CAVGA\\
		& \cite{sabokrou2018avid} & \cite{bergmann2018improving} &\cite{bergmann2018improving} & \cite{schlegl2017unsupervised} &\cite{dehaene2020iterative} &\cite{abati2019latent}     &-D$_{u}$ &-R$_{u}$ &-D$_{w}$ &-R$_{w}$\\
\midrule
		Bottle &\cellcolor{blue!40}0.88    &\cellcolor{blue!40}0.88  &\cellcolor{blue!20}0.80 &\cellcolor{blue!10}0.69 &\cellcolor{blue!30}0.86 &\cellcolor{blue!30}0.86 &\cellcolor{blue!50}0.89 &\cellcolor{blue!60}0.91 &\cellcolor{blue!70}0.93 &\cellcolor{blue!80}\color{white}0.96\\

		Hazelnut &\cellcolor{blue!40}0.86 &\cellcolor{blue!15}0.54  &\cellcolor{blue!60}0.88 &\cellcolor{blue!10}0.50 &\cellcolor{blue!20}0.74 &\cellcolor{blue!25}0.80   &\cellcolor{blue!30}0.84  &\cellcolor{blue!50}0.87  &\cellcolor{blue!70}0.90 &\cellcolor{blue!80}\color{white}0.92\\
		          
		Capsule  &\cellcolor{blue!40}0.85 &\cellcolor{blue!15}0.61  &\cellcolor{blue!20}0.62 &\cellcolor{blue!10}0.58 &\cellcolor{blue!50}0.86 &\cellcolor{blue!25}0.71  &\cellcolor{blue!30}0.83 &\cellcolor{blue!60}0.87  &\cellcolor{blue!70}0.89 &\cellcolor{blue!80}\color{white}0.93\\
		          
		Metal Nut &\cellcolor{blue!20}0.63 &\cellcolor{blue!5}0.54  &\cellcolor{blue!50}0.73 &\cellcolor{blue!0}0.50 &\cellcolor{blue!60}0.78 &\cellcolor{blue!30}0.67  &\cellcolor{blue!30}0.67 &\cellcolor{blue!40}0.71   &\cellcolor{blue!70}0.81 &\cellcolor{blue!80}\color{white}0.88\\
		          
		Leather  &\cellcolor{blue!30}0.58 &\cellcolor{blue!20}0.46  &\cellcolor{blue!5}0.44 &\cellcolor{blue!15}0.52 &\cellcolor{blue!50}0.71 &\cellcolor{blue!40}0.70   &\cellcolor{blue!50}0.71 &\cellcolor{blue!60}0.75  &\cellcolor{blue!70}0.80 &\cellcolor{blue!80}\color{white}0.84\\
		          
		Pill  &\cellcolor{blue!40}0.86    &\cellcolor{blue!10}0.60  &\cellcolor{blue!20}0.62 &\cellcolor{blue!20}0.62 &\cellcolor{blue!25}0.80 &\cellcolor{blue!30}0.85   &\cellcolor{blue!50}0.88 &\cellcolor{blue!60}0.91  &\cellcolor{blue!70}0.93 &\cellcolor{blue!80}\color{white}0.97\\
		          
		Wood    &\cellcolor{blue!40}0.83  &\cellcolor{blue!40}0.83  &\cellcolor{blue!20}0.74 &\cellcolor{blue!10}0.68 &\cellcolor{blue!80}\color{white}0.89 &\cellcolor{blue!30}0.75  &\cellcolor{blue!60}0.85  &\cellcolor{blue!70}0.88  &\cellcolor{blue!80}\color{white}0.89&\cellcolor{blue!80}\color{white}0.89\\
		          
		Carpet &\cellcolor{blue!30}0.70    &\cellcolor{blue!20}0.67  &\cellcolor{blue!15}0.50 &\cellcolor{blue!10}0.49 &\cellcolor{blue!20}0.67 &\cellcolor{blue!50}0.74  &\cellcolor{blue!40}0.73  &\cellcolor{blue!60}0.78  &\cellcolor{blue!70}0.80 &\cellcolor{blue!80}\color{white}0.82\\
		          
		Tile   &\cellcolor{blue!20}0.66   &\cellcolor{blue!15}0.52  &\cellcolor{blue!60}0.77 &\cellcolor{blue!10}0.51 &\cellcolor{blue!70}0.81 &\cellcolor{blue!30}0.70  &\cellcolor{blue!30}0.70  &\cellcolor{blue!50}0.72  &\cellcolor{blue!70}0.81 &\cellcolor{blue!80}\color{white}0.86\\
		          
		Grid   &\cellcolor{blue!20}0.59   &\cellcolor{blue!30}0.69  &\cellcolor{blue!50}0.78 &\cellcolor{blue!10}0.51 &\cellcolor{blue!80}\color{white}0.83 &\cellcolor{blue!15}0.54   &\cellcolor{blue!40}0.75  &\cellcolor{blue!50}0.78  &\cellcolor{blue!60}0.79&\cellcolor{blue!70}0.81\\
		          
		Cable &\cellcolor{blue!50}0.64    &\cellcolor{blue!30}0.61  &\cellcolor{blue!20}0.56 &\cellcolor{blue!10}0.53 &\cellcolor{blue!20}0.56 &\cellcolor{blue!30}0.61    &\cellcolor{blue!40}0.63  &\cellcolor{blue!60}0.67 &\cellcolor{blue!70}0.86 &\cellcolor{blue!80}\color{white}0.97\\
		          
		Transistor &\cellcolor{blue!20}0.58 &\cellcolor{blue!15}0.52  &\cellcolor{blue!40}0.71 &\cellcolor{blue!25}0.67 &\cellcolor{blue!30}0.70 &\cellcolor{blue!10}0.50  &\cellcolor{blue!50}0.73  &\cellcolor{blue!60}0.75  &\cellcolor{blue!70}0.80&\cellcolor{blue!80}\color{white}0.89\\
		          
		Toothbrush  &\cellcolor{blue!10}0.73 &\cellcolor{blue!15}0.74  &\cellcolor{blue!70}0.98 &\cellcolor{blue!5}0.57 &\cellcolor{blue!30}0.89 &\cellcolor{blue!30}0.89  &\cellcolor{blue!40}0.91 &\cellcolor{blue!60}0.97  &\cellcolor{blue!50}0.96 &\cellcolor{blue!80}\color{white}0.99\\
		          
		Screw &\cellcolor{blue!25}0.66    &\cellcolor{blue!15}0.51  &\cellcolor{blue!30}0.69 &\cellcolor{blue!10}0.35 &\cellcolor{blue!40}0.71 &\cellcolor{blue!50}0.75 &\cellcolor{blue!60}0.77 &\cellcolor{blue!70}0.78  &\cellcolor{blue!80}\color{white}0.79 &\cellcolor{blue!80}\color{white}0.79\\
		          
		Zipper  &\cellcolor{blue!30}0.84  &\cellcolor{blue!20}0.80  &\cellcolor{blue!20}0.80 &\cellcolor{blue!10}0.59 &\cellcolor{blue!15}0.67 &\cellcolor{blue!50}0.88   &\cellcolor{blue!40}0.87  &\cellcolor{blue!60}0.94  &\cellcolor{blue!70}0.95 &\cellcolor{blue!80}\color{white}0.96\\
\midrule
mean &\cellcolor{blue!30}0.73 &\cellcolor{blue!15}0.63 &\cellcolor{blue!20}0.71 &\cellcolor{blue!5}0.55 &\cellcolor{blue!40}0.77 &\cellcolor{blue!30}0.73  &\cellcolor{blue!50}0.78 &\cellcolor{blue!60}0.82 &\cellcolor{blue!70}0.86 &\cellcolor{blue!80}\color{white}0.90\\
\bottomrule
\end{tabular}

\end{center}

\end{table}
Fig. \ref{qual_mvtec} (b) illustrates the qualitative results and Table \ref{table_iou_st} shows that CAVGA also outperforms the baseline methods in mean IoU and mean AuROC on the mSTC dataset. Table \ref{table_lag} shows that CAVGA outperforms the most competitive baseline Patho-GAN \cite{wang2019pathology} by 16\% in IoU on the LAG dataset. CAVGA is practically reasonable to train on a single GTX 1080Ti GPU, having comparable training and testing time with baseline methods. 

\textbf{Performance on anomaly detection:} Table \ref{table_accu_mvtec} shows that CAVGA$_{u}$ outperforms the baselines in the mean of accuracy of correctly classified anomalous images and normal images on the MVTAD dataset. CAVGA-D$_{u}$ outperforms the best performing baseline ($\gamma$-VAE$_{\text{g}}$) in mean of classification accuracy by 1.3\%. Table \ref{table_iou_st} and Table \ref{table_lag} show that CAVGA outperforms the baseline methods in classification accuracy on both the mSTC and LAG datasets by 2.6\% and 1.1\% respectively. Furthermore,  Table \ref{table_novelty_detection} shows that CAVGA-D$_{u}$ outperforms all the baselines in mean AuROC in the unsupervised setting on the MNIST, CIFAR-10 and Fashion-MNIST datasets. CAVGA-D$_{u}$ also outperforms MemAE \cite{gong2019memorizing} and $\beta$-VAE \cite{higgins2017beta} by 1.1\% and 8\% on MNIST and by 21\% and 38\% on CIFAR-10 datasets respectively. CAVGA-D$_{u}$ also outperforms all the listed baselines in mean AuROC on the Fashion-MNIST dataset.

\begin{table}[t!]
\begin{center}
\caption{
Performance comparison of anomaly detection in terms of AuROC and mean AuROC with the SOTA methods on MNIST ($D_M$) and CIFAR-10 ($D_C$) datasets . We also report the mean AuROC on Fashion-MNIST ($D_F$) dataset 
}
\label{table_novelty_detection}
\scriptsize
\begin{tabular}{ccccccccccc}
\toprule
		 Dataset &Class & $\gamma$-VAE$_{\text{g}}$ &LSA  &OCGAN  &ULSLM  &CapsNet$_\text{PP}$  &CapsNet$_\text{RE}$  &AnoGAN & ADGAN   &CAVGA \\
		  & &\cite{dehaene2020iterative} & \cite{abati2019latent} & \cite{perera2019ocgan} & \cite{wolf2020unsupervised} & \cite{li2019exploring} & \cite{li2019exploring} & \cite{schlegl2017unsupervised} &  \cite{deecke2018image} & -D$_{u}$\\
\midrule
            	            &0 &\cellcolor{blue!40}0.991 &\cellcolor{blue!50}0.993 &\cellcolor{blue!70}0.998 &\cellcolor{blue!40}0.991    &\cellcolor{blue!70}0.998     &\cellcolor{blue!20}0.947      &\cellcolor{blue!30}0.990    &\cellcolor{blue!80}\color{white}0.999    &\cellcolor{blue!60}0.994\\
            	            
            	            &1  &\cellcolor{blue!40}0.996 &\cellcolor{blue!80}\color{white}0.999 &\cellcolor{blue!80}\color{white}0.999  &\cellcolor{blue!15}0.972   &\cellcolor{blue!20}0.990     &\cellcolor{blue!10}0.907      &\cellcolor{blue!60}0.998    &\cellcolor{blue!30}0.992     &\cellcolor{blue!50}0.997\\
            	            
            	            &2 &\cellcolor{blue!60}0.983 &\cellcolor{blue!30}0.959 &\cellcolor{blue!15}0.942  &\cellcolor{blue!10}0.919     &\cellcolor{blue!70}0.984     &\cellcolor{blue!50}0.970      &\cellcolor{blue!5}0.888    &\cellcolor{blue!40}0.968                &\cellcolor{blue!80}\color{white}0.989\\
            	            
            	            &3  &\cellcolor{blue!70}0.978 &\cellcolor{blue!50}0.966  &\cellcolor{blue!40}0.963  &\cellcolor{blue!15}0.943  &\cellcolor{blue!60}0.976     &\cellcolor{blue!25}0.949      &\cellcolor{blue!5}0.913    &\cellcolor{blue!30}0.953        &\cellcolor{blue!80}\color{white}0.983\\
            	            
            	            &4  &\cellcolor{blue!70}0.976 &\cellcolor{blue!40}0.956 &\cellcolor{blue!60}0.975   &\cellcolor{blue!20}0.942  &\cellcolor{blue!15}0.935     &\cellcolor{blue!5}0.872      &\cellcolor{blue!30}0.944    &\cellcolor{blue!50}0.960        &\cellcolor{blue!80}\color{white}0.977\\
            	            
	         $D_M$ \cite{lecun1998gradient}  &5 &\cellcolor{blue!70}0.972    &\cellcolor{blue!30}0.964 
	         &\cellcolor{blue!80} \color{white}0.980  &\cellcolor{blue!5}0.872  &\cellcolor{blue!60}0.970     &\cellcolor{blue!40}0.966      &\cellcolor{blue!15}0.912    &\cellcolor{blue!20}0.955      &\cellcolor{blue!50}0.968\\
	         
            	            &6 &\cellcolor{blue!70}0.993 &\cellcolor{blue!80}\color{white}0.994 &\cellcolor{blue!60}0.991 &\cellcolor{blue!50}0.988      &\cellcolor{blue!30}0.942    &\cellcolor{blue!10}0.909 &\cellcolor{blue!20}0.925    &\cellcolor{blue!40}0.980    &\cellcolor{blue!50}0.988\\
            	            
            	            &7 &\cellcolor{blue!60}0.981 &\cellcolor{blue!50}0.980 &\cellcolor{blue!60}0.981  &\cellcolor{blue!20}0.939     &\cellcolor{blue!80}\color{white}0.987 &\cellcolor{blue!15}0.934      &\cellcolor{blue!40}0.964    &\cellcolor{blue!30}0.950       &\cellcolor{blue!70}0.986\\
            	            
            	            &8 &\cellcolor{blue!60}0.980
            	            &\cellcolor{blue!30}0.953 &\cellcolor{blue!20}0.939   &\cellcolor{blue!50}0.960    &\cellcolor{blue!80}\color{white}0.993     &\cellcolor{blue!15}0.929      &\cellcolor{blue!5}0.883    &\cellcolor{blue!40}0.959    &\cellcolor{blue!70}0.988\\
            	            
            	            &9  &\cellcolor{blue!50}0.967 &\cellcolor{blue!60}0.981 &\cellcolor{blue!60}0.981  &\cellcolor{blue!50}0.967     &\cellcolor{blue!70}0.990     &\cellcolor{blue!10}0.871      &\cellcolor{blue!30}0.958    &\cellcolor{blue!40}0.965     &\cellcolor{blue!80}\color{white}0.991\\
            	            
            	            \midrule
            	            &mean &\cellcolor{blue!70}0.982  &\cellcolor{blue!50}0.975 &\cellcolor{blue!50}0.975 &\cellcolor{blue!30}0.949 &\cellcolor{blue!60}0.977     &\cellcolor{blue!10}0.925      &\cellcolor{blue!20}0.937    &\cellcolor{blue!40}0.968    &\cellcolor{blue!80}\color{white}0.986\\
            	            
            	            \midrule[.75pt]
                            &0 &\cellcolor{blue!50}0.702 &\cellcolor{blue!70}0.735  &\cellcolor{blue!80}\color{white}0.757 
                            &\cellcolor{blue!60}0.740     &\cellcolor{blue!25}0.622     &\cellcolor{blue!15}0.371      &\cellcolor{blue!20}0.610    &\cellcolor{blue!40}0.661    &\cellcolor{blue!30}0.653\\
                            
            	            &1 &\cellcolor{blue!60}0.663 &\cellcolor{blue!50}0.580  &\cellcolor{blue!30}0.531  &\cellcolor{blue!70}0.747   &\cellcolor{blue!20}0.455     &\cellcolor{blue!60}0.737      &\cellcolor{blue!40}0.565    &\cellcolor{blue!10}0.435     &\cellcolor{blue!80}\color{white}0.784\\
            	            
            	            &2  &\cellcolor{blue!60}0.680 &\cellcolor{blue!70}0.690 &\cellcolor{blue!30}0.640  &\cellcolor{blue!20}0.628    &\cellcolor{blue!50}0.671     &\cellcolor{blue!15}0.421      &\cellcolor{blue!40}0.648    &\cellcolor{blue!25}0.636    &\cellcolor{blue!80}\color{white}0.761\\
            	            
            	            &3  &\cellcolor{blue!70}0.713 &\cellcolor{blue!20}0.542 &\cellcolor{blue!50}0.620 
            	            &\cellcolor{blue!30}0.572    &\cellcolor{blue!60}0.675     &\cellcolor{blue!40}0.588      &\cellcolor{blue!10}0.528    &\cellcolor{blue!5}0.488    &\cellcolor{blue!80}\color{white}0.747\\
            	            
            	            &4 &\cellcolor{blue!60}0.770 &\cellcolor{blue!50}0.761 &\cellcolor{blue!40}0.723  &\cellcolor{blue!25}0.678   &\cellcolor{blue!30}0.683     &\cellcolor{blue!15}0.388      &\cellcolor{blue!20}0.670    &\cellcolor{blue!80}\color{white}0.794    &\cellcolor{blue!70}0.775\\
            	            
	  $D_C$ \cite{krizhevsky2009learning}          &5 &\cellcolor{blue!80}\color{white}0.689 &\cellcolor{blue!5}0.546  &\cellcolor{blue!50}0.620 &\cellcolor{blue!40}0.602     &\cellcolor{blue!60}0.635     &\cellcolor{blue!30}0.601      &\cellcolor{blue!20}0.592    &\cellcolor{blue!70}0.640        &\cellcolor{blue!10}0.552\\
                            &6 &\cellcolor{blue!70}0.805 &\cellcolor{blue!50}0.751  &\cellcolor{blue!30}0.723 &\cellcolor{blue!60}0.753    &\cellcolor{blue!40}0.727     &\cellcolor{blue!5}0.491      &\cellcolor{blue!20}0.625    &\cellcolor{blue!25}0.685       &\cellcolor{blue!80}\color{white}0.813\\
                            
            	            &7 &\cellcolor{blue!40}0.588 &\cellcolor{blue!5}0.535  &\cellcolor{blue!20}0.575  &\cellcolor{blue!70}0.685    &\cellcolor{blue!60}0.673     &\cellcolor{blue!50}0.631      &\cellcolor{blue!30}0.576    &\cellcolor{blue!15}0.559      &\cellcolor{blue!80}\color{white}0.745\\
            	            
            	            &8  &\cellcolor{blue!70}0.813  &\cellcolor{blue!25}0.717  &\cellcolor{blue!80}\color{white}0.820  &\cellcolor{blue!40}0.781    &\cellcolor{blue!20}0.710     &\cellcolor{blue!15}0.410      &\cellcolor{blue!30}0.723    &\cellcolor{blue!50}0.798       &\cellcolor{blue!60}0.801\\
            	            
            	            &9 &\cellcolor{blue!70}0.744 &\cellcolor{blue!15}0.548 &\cellcolor{blue!20}0.554  &\cellcolor{blue!80}\color{white}0.795    &\cellcolor{blue!10}0.466     &\cellcolor{blue!50}0.671      &\cellcolor{blue!30}0.582    &\cellcolor{blue!40}0.643       &\cellcolor{blue!60}0.741\\
            	            
            	            \midrule
            	            &mean &\cellcolor{blue!60}0.717 &\cellcolor{blue!40}0.641  &\cellcolor{blue!50}0.656 &\cellcolor{blue!70}0.736   &\cellcolor{blue!20}0.612     &\cellcolor{blue!10}0.531      &\cellcolor{blue!20}0.612    &\cellcolor{blue!30}0.634     &\cellcolor{blue!80}\color{white}0.737\\
            	            \midrule[.75pt]
    $D_F$ \cite{xiao2017fashion}    &mean  &\cellcolor{blue!40}0.873 &\cellcolor{blue!60}0.876 &\cellcolor{blue!10}-  &\cellcolor{blue!10}- &\cellcolor{blue!30}0.765     &\cellcolor{blue!10}0.679      &\cellcolor{blue!10}-    &\cellcolor{blue!10}-      &\cellcolor{blue!80}\color{white}0.885\\
            	            
    \bottomrule 
\end{tabular}
\end{center}
\end{table}

\section{Ablation Study}
\label{ablation}

\begin{table}[h!]
\begin{center}
\renewcommand{\arraystretch}{.8}
\setlength{\tabcolsep}{2.0pt}
\caption{
The ablation study on 5 randomly chosen categories showing anomaly localization in IoU on the MVTAD dataset. The mean of all 15 categories is reported. CAVGA-R$_{u}^{*}$ and CAVGA-R$_{w}^{*}$ are our base architecture with a flattened $z$ in the unsupervised and weakly supervised settings respectively. ``conv $z$" means using convolutional $z$
}
\label{ablation_table}
\scriptsize
\begin{tabular}{@{}ccccc@{}cccc@{}}
\toprule
		  Method &CAVGA &CAVGA &CAVGA &CAVGA &CAVGA &CAVGA &CAVGA &CAVGA\\
		              &-R$_{u}^{*}$  &-R$_{u}^{*}$ &-R$_{u}$  &-R$_{u}$  &-R$_{w}^{*}$ &-R$_{w}^{*}$ &-R$_{w}$    &-R$_{w}$\\
		  &                & $+$ $L_{ae}$      & $+$ conv $z$ & $+$ conv $z$  &               & $+$ $L_{cga}$        & $+$ conv $z$     & $+$ conv $z$\\
		  Category &                &      &  & $+$ $L_{ae}$  &               &         &   & $+$ $L_{cga}$\\
\midrule
		Column ID &$c_1$ &$c_2$ &$c_3$ &$c_4$ &$c_5$ &$c_6$ &$c_7$ &$c_8$\\
\midrule
    	Bottle    &\cellcolor{blue!10}0.24 &\cellcolor{blue!50}0.27 &\cellcolor{blue!30}0.26  &\cellcolor{blue!80}\color{white}0.33 &\cellcolor{blue!10}0.16  &\cellcolor{blue!50}0.34 &\cellcolor{blue!30}0.28   &\cellcolor{blue!80}\color{white}0.39   \\
    	Hazelnut  &\cellcolor{blue!10}0.16 &\cellcolor{blue!30}0.26 &\cellcolor{blue!50}0.31  &\cellcolor{blue!80}\color{white}0.47 &  \cellcolor{blue!10}0.51   &\cellcolor{blue!50}0.76 &\cellcolor{blue!30}0.67   &\cellcolor{blue!80}\color{white}0.79   \\
    	Capsule   &\cellcolor{blue!10}0.09  &\cellcolor{blue!50}0.22   &\cellcolor{blue!30}0.14  &\cellcolor{blue!80}\color{white}0.31  &\cellcolor{blue!10}0.18  &\cellcolor{blue!50}0.36   &\cellcolor{blue!30}0.27 &\cellcolor{blue!80}\color{white}0.41\\
		
		Metal Nut &\cellcolor{blue!10}0.28 &\cellcolor{blue!50}0.38 &\cellcolor{blue!30}0.34  &\cellcolor{blue!80}\color{white}0.45 &  \cellcolor{blue!10}0.25   &\cellcolor{blue!50}0.38 &\cellcolor{blue!30}0.28   &\cellcolor{blue!80}\color{white}0.46   \\
		Leather   &\cellcolor{blue!10}0.55  &\cellcolor{blue!50}0.71   &\cellcolor{blue!30}0.64  &\cellcolor{blue!80}\color{white}0.79  &\cellcolor{blue!10}0.72  &\cellcolor{blue!50}0.79   &\cellcolor{blue!30}0.75 &\cellcolor{blue!80}\color{white}0.84\\
\midrule
mean &\cellcolor{blue!10}0.24 &\cellcolor{blue!50}0.34 &\cellcolor{blue!30}0.33 &\cellcolor{blue!80}\color{white}0.47 &\cellcolor{blue!10}0.39 &\cellcolor{blue!50}0.52 &\cellcolor{blue!30}0.48  &\cellcolor{blue!80}\color{white}0.60\\
\bottomrule 
\end{tabular}
\end{center}
\end{table}

\begin{figure}
\centering
\includegraphics[width= .8 \linewidth]{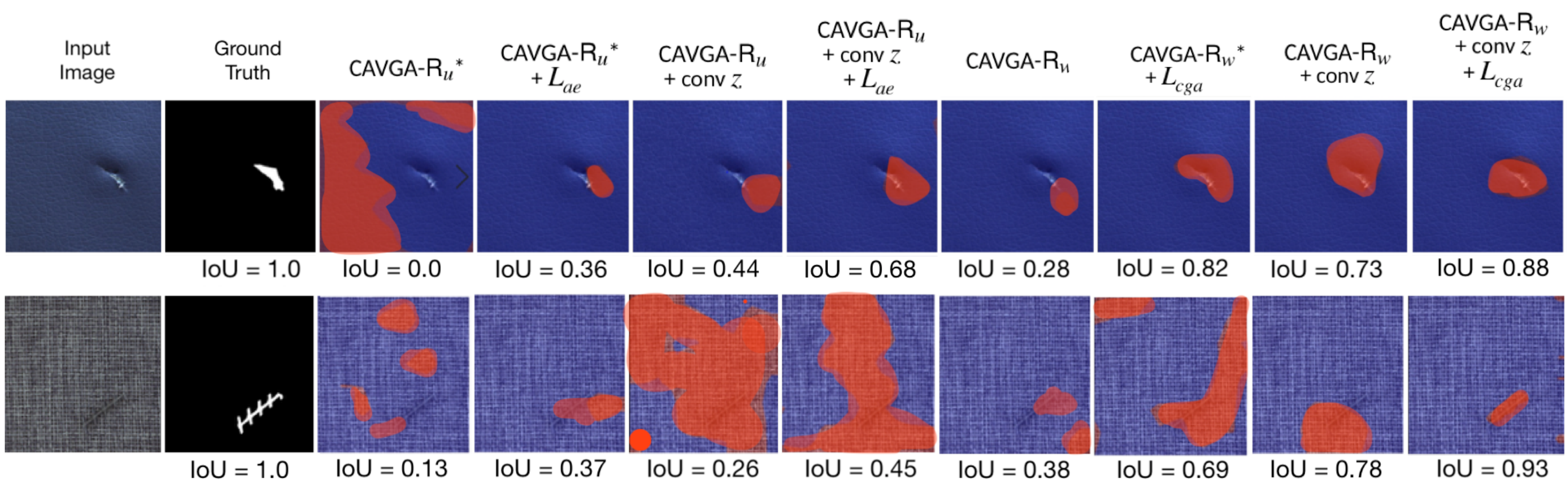}
\caption{Qualitative results of the ablation study to illustrate the performance of the anomaly localization on the MVTAD dataset}
\label{fig_ablation}

\end{figure}

All the ablation studies are performed on 15 categories on the MVTAD dataset, of which 5 are reported here. The mean of all 15 categories is shown in Table \ref{ablation_table}. We illustrate the effectiveness of the convolutional $z$ in CAVGA, $L_{ae}$ in the unsupervised setting, and $L_{cga}$ in the weakly supervised setting. The qualitative results are shown in Fig. \ref{fig_ablation}. The column IDs to refer to the columns in Table \ref{ablation_table}.

\textbf{Effect of convolutional latent variable $z$:} To show the effectiveness of the convolutional $z$, we flatten the output of the encoder of CAVGA-R$_{u}$ and CAVGA-R$_{w}$, and connect it to a fully connected layer as latent variable. Following \cite{bergmann2018improving}, the dimension of latent variable is chosen as $100$. We call these network as CAVGA-R$_{u}^{*}$ and CAVGA-R$_{w}^{*}$ in the unsupervised and weakly supervised settings respectively. In the unsupervised setting, we train CAVGA-R$_{u}$ and CAVGA-R$_{u}^{*}$  using $L + L_{adv}$ as our objective function and compute the anomalous attention map from the feature map of the latent variable during inference. Similarly, in the weakly supervised setting, we train CAVGA-R$_{w}$ and CAVGA-R$_{w}^{*}$  using $L + L_{adv} + L_{bce}$ as our objective function and compute the anomalous attention map from the classifier's prediction during inference. Comparing column $c_1$ with $c_3$ and $c_5$ with $c_7$ in Table \ref{ablation_table}, we observe that preserving the spatial relation of the input and latent variable through the convolutional $z$ improves the IoU in anomaly localization without the use of $L_{ae}$ in the unsupervised setting and $L_{cga}$ in the weakly supervised setting. Furthermore, comparing column $c_2$ with $c_4$ and $c_6$ with $c_8$ in Table \ref{ablation_table}, we observe that using convolutional $z$ in CAVGA-R$_{u}$ and CAVGA-R$_{w}$ outperforms using a flattened latent variable even with the help of $L_{ae}$ in the unsupervised setting and $L_{cga}$ in the weakly supervised setting.

\textbf{Effect of attention expansion loss $L_{ae}$:} To test the effectiveness of using $L_{ae}$ in the unsupervised setting, we train CAVGA-R$_{u}^{*}$ and CAVGA-R$_{u}$ with eq. \ref{eqn_finalunsuper}. During inference, the anomalous attention map is computed to localize the anomaly. Comparing column $c_1$ with $c_2$ and $c_3$ with $c_4$ in Table \ref{ablation_table}, we observe that $L_{ae}$ enhances the IoU regardless of a flattened or convolutional latent variable.

\textbf{Effect of complementary guided attention loss $L_{cga}$:} We show the effectiveness of $L_{cga}$ by training CAVGA-R$_{w}^{*}$ and CAVGA-R$_w$ using eq. \ref{eqn_weaksuper}. Comparing column $c_5$ with $c_6$ and $c_7$ with $c_8$ in Table \ref{ablation_table}, we find that using $L_{cga}$ enhances the IoU regardless of a flattened or convolutional latent variable.

\section{Conclusion}
\label{conclusion}
We propose an end-to-end convolutional adversarial variational autoencoder using guided attention which is a novel use of this technique for anomaly localization. Applicable to different network architectures, our attention expansion loss and complementary guided attention loss improve the performance of anomaly localization in the unsupervised and weakly supervised (with only 2\% extra anomalous images for training) settings respectively. We quantitatively and qualitatively show that CAVGA outperforms the state-of-the-art (SOTA) anomaly localization methods on the MVTAD, mSTC and LAG datasets. We also show CAVGA's ability to outperform SOTA anomaly detection methods on the MVTAD, mSTC, LAG, MNIST, Fashion-MNIST and CIFAR-10 datasets. \\

\textbf{Acknowledgments :} This work was done when Shashanka was an intern and Kuan-Chuan was a Staff Scientist at Siemens. Shashanka's effort was partially supported by DARPA under Grant D19AP00032.  
\clearpage

\bibliographystyle{splncs04}
\bibliography{egbib}

\begin{thebibliography}{10}
\providecommand{\url}[1]{\texttt{#1}}
\providecommand{\urlprefix}{URL }
\providecommand{\doi}[1]{https://doi.org/#1}

\bibitem{iclr20_code}
Code for iterative energy-based projection on a normal data manifold for
  anomaly localization.
  \url{https://qiita.com/kogepan102/items/122b2862ad5a51180656}, accessed on:
  2020-02-29

\bibitem{abati2019latent}
Abati, D., Porrello, A., Calderara, S., Cucchiara, R.: Latent space
  autoregression for novelty detection. In: Proceedings of the IEEE Conference
  on Computer Vision and Pattern Recognition. pp. 481--490 (2019)

\bibitem{akcay2018ganomaly}
Akcay, S., Atapour-Abarghouei, A., Breckon, T.P.: {GAN}omaly: Semi-supervised
  anomaly detection via adversarial training. In: Asian Conference on Computer
  Vision. pp. 622--637. Springer (2018)

\bibitem{baur2018deep}
Baur, C., Wiestler, B., Albarqouni, S., Navab, N.: Deep autoencoding models for
  unsupervised anomaly segmentation in brain mr images. In: International
  MICCAI Brainlesion Workshop. pp. 161--169. Springer (2018)

\bibitem{bergmann2019mvtec}
Bergmann, P., Fauser, M., Sattlegger, D., Steger, C.: {MVTec AD}--a
  comprehensive real-world dataset for unsupervised anomaly detection. In:
  Proceedings of the IEEE Conference on Computer Vision and Pattern
  Recognition. pp. 9592--9600 (2019)

\bibitem{bergmann2018improving}
Bergmann, P., L{\"o}we, S., Fauser, M., Sattlegger, D., Steger, C.: Improving
  unsupervised defect segmentation by applying structural similarity to
  autoencoders. In: International Joint Conference on Computer Vision, Imaging
  and Computer Graphics Theory and Applications (VISIGRAPP). vol.~5 (2019)

\bibitem{bian2019novel}
Bian, J., Hui, X., Sun, S., Zhao, X., Tan, M.: A novel and efficient
  cvae-gan-based approach with informative manifold for semi-supervised anomaly
  detection. IEEE Access  \textbf{7},  88903--88916 (2019)

\bibitem{bottger2016real}
B{\"o}ttger, T., Ulrich, M.: Real-time texture error detection on textured
  surfaces with compressed sensing. Pattern Recognition and Image Analysis
  \textbf{26}(1),  88--94 (2016)

\bibitem{brock2018large}
Brock, A., Donahue, J., Simonyan, K.: Large scale {GAN} training for high
  fidelity natural image synthesis. In: International Conference on Learning
  Representations (2019)

\bibitem{cheng2013abnormal}
Cheng, K.W., Chen, Y.T., Fang, W.H.: Abnormal crowd behavior detection and
  localization using maximum sub-sequence search. In: Proceedings of the 4th
  ACM/IEEE international workshop on Analysis and retrieval of tracked events
  and motion in imagery stream. pp. 49--58. ACM (2013)

\bibitem{daniel2019deep}
Daniel, T., Kurutach, T., Tamar, A.: Deep variational semi-supervised novelty
  detection. arXiv preprint arXiv:1911.04971  (2019)

\bibitem{deecke2018image}
Deecke, L., Vandermeulen, R., Ruff, L., Mandt, S., Kloft, M.: Image anomaly
  detection with generative adversarial networks. In: Joint European Conference
  on Machine Learning and Knowledge Discovery in Databases. pp. 3--17. Springer
  (2018)

\bibitem{dehaene2020iterative}
Dehaene, D., Frigo, O., Combrexelle, S., Eline, P.: Iterative energy-based
  projection on a normal data manifold for anomaly localization. International
  Conference on Learning Representations  (2020)

\bibitem{dieng2019avoiding}
Dieng, A.B., Kim, Y., Rush, A.M., Blei, D.M.: Avoiding latent variable collapse
  with generative skip models. In: The 22nd International Conference on
  Artificial Intelligence and Statistics. pp. 2397--2405 (2019)

\bibitem{dimokranitou2017adversarial}
Dimokranitou, A.: Adversarial autoencoders for anomalous event detection in
  images. Ph.D. thesis (2017)

\bibitem{gong2019memorizing}
Gong, D., Liu, L., Le, V., Saha, B., Mansour, M.R., Venkatesh, S., Hengel,
  A.v.d.: Memorizing normality to detect anomaly: Memory-augmented deep
  autoencoder for unsupervised anomaly detection. In: Proceedings of the IEEE
  International Conference on Computer Vision. pp. 1705--1714 (2019)

\bibitem{goodfellow2014generative}
Goodfellow, I., Pouget-Abadie, J., Mirza, M., Xu, B., Warde-Farley, D., Ozair,
  S., Courville, A., Bengio, Y.: Generative adversarial nets. In: Advances in
  neural information processing systems. pp. 2672--2680 (2014)

\bibitem{gutoski2017detection}
Gutoski, M., Aquino, N.M.R., Ribeiro, M., Lazzaretti, E., Lopes, S.: Detection
  of video anomalies using convolutional autoencoders and one-class support
  vector machines. In: XIII Brazilian Congress on Computational Intelligence,
  2017 (2017)

\bibitem{he2016deep}
He, K., Zhang, X., Ren, S., Sun, J.: Deep residual learning for image
  recognition. In: Proceedings of the IEEE Conference on Computer Vision and
  Pattern Recognition. pp. 770--778 (2016)

\bibitem{hendrycks2018deep}
Hendrycks, D., Mazeika, M., Dietterich, T.G.: Deep anomaly detection with
  outlier exposure. In: International Conference on Learning Representations
  (2019)

\bibitem{higgins2017beta}
Higgins, I., Matthey, L., Pal, A., Burgess, C., Glorot, X., Botvinick, M.,
  Mohamed, S., Lerchner, A.: beta-{VAE}: Learning basic visual concepts with a
  constrained variational framework. International Conference on Learning
  Representations  \textbf{2}(5), ~6 (2017)

\bibitem{Kimura_2020_WACV}
Kimura, D., Chaudhury, S., Narita, M., Munawar, A., Tachibana, R.: Adversarial
  discriminative attention for robust anomaly detection. In: The IEEE Winter
  Conference on Applications of Computer Vision (WACV) (March 2020)

\bibitem{kingma2013auto}
Kingma, D.P., Welling, M.: Auto-encoding variational bayes. In: International
  Conference on Learning Representations (2014)

\bibitem{kiran2018overview}
Kiran, B., Thomas, D., Parakkal, R.: An overview of deep learning based methods
  for unsupervised and semi-supervised anomaly detection in videos. Journal of
  Imaging  \textbf{4}(2), ~36 (2018)

\bibitem{krizhevsky2009learning}
Krizhevsky, A., Hinton, G., et~al.: Learning multiple layers of features from
  tiny images. Tech. rep., Citeseer (2009)

\bibitem{larsen2015autoencoding}
Larsen, A.B.L., S{\o}nderby, S.K., Larochelle, H., Winther, O.: Autoencoding
  beyond pixels using a learned similarity metric. In: International Conference
  on Machine Learning (2016)

\bibitem{lecun1998gradient}
LeCun, Y., Bottou, L., Bengio, Y., Haffner, P., et~al.: Gradient-based learning
  applied to document recognition. Proceedings of the IEEE  \textbf{86}(11),
  2278--2324 (1998)

\bibitem{li2018tell}
Li, K., Wu, Z., Peng, K.C., Ernst, J., Fu, Y.: Tell me where to look: Guided
  attention inference network. In: Proceedings of the IEEE Conference on
  Computer Vision and Pattern Recognition. pp. 9215--9223 (2018)

\bibitem{Li_2019_CVPR}
Li, L., Xu, M., Wang, X., Jiang, L., Liu, H.: Attention based glaucoma
  detection: A large-scale database and cnn model. In: The IEEE Conference on
  Computer Vision and Pattern Recognition (CVPR) (June 2019)

\bibitem{li2019exploring}
Li, X., Kiringa, I., Yeap, T., Zhu, X., Li, Y.: Exploring deep anomaly
  detection methods based on capsule net. International Conference on Machine
  Learning 2019 Workshop on Uncertainty and Robustness in Deep Learning  (2019)

\bibitem{liu2018future}
Liu, W., Luo, W., Lian, D., Gao, S.: Future frame prediction for anomaly
  detection--a new baseline. In: Proceedings of the IEEE Conference on Computer
  Vision and Pattern Recognition. pp. 6536--6545 (2018)

\bibitem{liu2019towards}
Liu, W., Li, R., Zheng, M., Karanam, S., Wu, Z., Bhanu, B., Radke, R.J., Camps,
  O.: Towards visually explaining variational autoencoders. Proceedings of the
  IEEE Conference on Computer Vision and Pattern Recognition  (2020)

\bibitem{liu2015faceattributes}
Liu, Z., Luo, P., Wang, X., Tang, X.: Deep learning face attributes in the
  wild. In: Proceedings of International Conference on Computer Vision (ICCV)
  (December 2015)

\bibitem{makhzani2015adversarial}
Makhzani, A., Shlens, J., Jaitly, N., Goodfellow, I., Frey, B.: Adversarial
  autoencoders. In: International Conference on Learning Representations (2016)

\bibitem{masana2018metric}
Masana, M., Ruiz, I., Serrat, J., van~de Weijer, J., Lopez, A.M.: Metric
  learning for novelty and anomaly detection. In: British Machine Vision
  Conference (BMVC) (2018)

\bibitem{matteoli2014overview}
Matteoli, S., Diani, M., Theiler, J.: An overview of background modeling for
  detection of targets and anomalies in hyperspectral remotely sensed imagery.
  IEEE Journal of Selected Topics in Applied Earth Observations and Remote
  Sensing  \textbf{7}(6),  2317--2336 (2014)

\bibitem{napoletano2018anomaly}
Napoletano, P., Piccoli, F., Schettini, R.: Anomaly detection in nanofibrous
  materials by {CNN}-based self-similarity. Sensors  \textbf{18}(1), ~209
  (2018)

\bibitem{nguyen2018weakly}
Nguyen, P., Liu, T., Prasad, G., Han, B.: Weakly supervised action localization
  by sparse temporal pooling network. In: Proceedings of the IEEE Conference on
  Computer Vision and Pattern Recognition. pp. 6752--6761 (2018)

\bibitem{oquab2015object}
Oquab, M., Bottou, L., Laptev, I., Sivic, J.: Is object localization for
  free?-weakly-supervised learning with convolutional neural networks. In:
  Proceedings of the IEEE Conference on Computer Vision and Pattern
  Recognition. pp. 685--694 (2015)

\bibitem{pawlowski2018unsupervised}
Pawlowski, N., Lee, M.C., Rajchl, M., McDonagh, S., Ferrante, E., Kamnitsas,
  K., Cooke, S., Stevenson, S., Khetani, A., Newman, T., et~al.: Unsupervised
  lesion detection in brain {CT} using bayesian convolutional autoencoders. In:
  Medical Imaging with Deep Learning (2018)

\bibitem{perera2019ocgan}
Perera, P., Nallapati, R., Xiang, B.: {OCGAN}: One-class novelty detection
  using {GAN}s with constrained latent representations. In: Proceedings of the
  IEEE Conference on Computer Vision and Pattern Recognition. pp. 2898--2906
  (2019)

\bibitem{radford2015unsupervised}
Radford, A., Metz, L., Chintala, S.: Unsupervised representation learning with
  deep convolutional generative adversarial networks. In: International
  Conference on Learning Representations (2016)

\bibitem{ravanbakhsh2019training}
Ravanbakhsh, M., Sangineto, E., Nabi, M., Sebe, N.: Training adversarial
  discriminators for cross-channel abnormal event detection in crowds. In: 2019
  IEEE Winter Conference on Applications of Computer Vision (WACV). pp.
  1896--1904. IEEE (2019)

\bibitem{ruff2019deep}
Ruff, L., Vandermeulen, R.A., G{\"o}rnitz, N., Binder, A., M{\"u}ller, E.,
  M{\"u}ller, K.R., Kloft, M.: Deep semi-supervised anomaly detection.
  International Conference on Learning Representations  (2020)

\bibitem{russakovsky2015imagenet}
Russakovsky, O., Deng, J., Su, H., Krause, J., Satheesh, S., Ma, S., Huang, Z.,
  Karpathy, A., Khosla, A., Bernstein, M., et~al.: {ImageNet} large scale
  visual recognition challenge. International journal of computer vision
  \textbf{115}(3),  211--252 (2015)

\bibitem{sabokrou2018adversarially}
Sabokrou, M., Khalooei, M., Fathy, M., Adeli, E.: Adversarially learned
  one-class classifier for novelty detection. In: Proceedings of the IEEE
  Conference on Computer Vision and Pattern Recognition. pp. 3379--3388 (2018)

\bibitem{sabokrou2018avid}
Sabokrou, M., Pourreza, M., Fayyaz, M., Entezari, R., Fathy, M., Gall, J.,
  Adeli, E.: Avid: Adversarial visual irregularity detection. In: Asian
  Conference on Computer Vision. pp. 488--505. Springer (2018)

\bibitem{schlegl2017unsupervised}
Schlegl, T., Seeb{\"o}ck, P., Waldstein, S.M., Schmidt-Erfurth, U., Langs, G.:
  Unsupervised anomaly detection with generative adversarial networks to guide
  marker discovery. In: International Conference on Information Processing in
  Medical Imaging. pp. 146--157. Springer (2017)

\bibitem{selvaraju2017grad}
Selvaraju, R.R., Cogswell, M., Das, A., Vedantam, R., Parikh, D., Batra, D.:
  Grad-cam: Visual explanations from deep networks via gradient-based
  localization. In: Proceedings of the IEEE International Conference on
  Computer Vision. pp. 618--626 (2017)

\bibitem{smilkov2017smoothgrad}
Smilkov, D., Thorat, N., Kim, B., Vi{\'e}gas, F., Wattenberg, M.: {SmoothGrad}:
  removing noise by adding noise. arXiv preprint arXiv:1706.03825  (2017)

\bibitem{springenberg2014striving}
Springenberg, J.T., Dosovitskiy, A., Brox, T., Riedmiller, M.: Striving for
  simplicity: The all convolutional net. arXiv preprint arXiv:1412.6806  (2014)

\bibitem{steger2001similarity}
Steger, C.: Similarity measures for occlusion, clutter, and illumination
  invariant object recognition. In: Joint Pattern Recognition Symposium. pp.
  148--154. Springer (2001)

\bibitem{vu2019anomaly}
Vu, H.S., Ueta, D., Hashimoto, K., Maeno, K., Pranata, S., Shen, S.M.: Anomaly
  detection with adversarial dual autoencoders. arXiv preprint arXiv:1902.06924
   (2019)

\bibitem{wang2019pathology}
Wang, X., Xu, M., Li, L., Wang, Z., Guan, Z.: Pathology-aware deep network
  visualization and its application in glaucoma image synthesis. In:
  International Conference on Medical Image Computing and Computer-Assisted
  Intervention. pp. 423--431. Springer (2019)

\bibitem{wang2019inductive}
Wang, Z., Fan, M., Muknahallipatna, S., Lan, C.: Inductive multi-view
  semi-supervised anomaly detection via probabilistic modeling. In: 2019 IEEE
  International Conference on Big Knowledge (ICBK). pp. 257--264. IEEE (2019)

\bibitem{wolf2020unsupervised}
Wolf, L., Benaim, S., Galanti, T.: Unsupervised learning of the set of local
  maxima. International Conference on Learning Representations  (2019)

\bibitem{xiao2017fashion}
Xiao, H., Rasul, K., Vollgraf, R.: Fashion-{MNIST}: a novel image dataset for
  benchmarking machine learning algorithms. arXiv preprint arXiv:1708.07747
  (2017)

\bibitem{zagoruyko2016paying}
Zagoruyko, S., Komodakis, N.: Paying more attention to attention: Improving the
  performance of convolutional neural networks via attention transfer. In:
  International Conference on Learning Representations (2017)

\bibitem{zenati2018efficient}
Zenati, H., Foo, C.S., Lecouat, B., Manek, G., Chandrasekhar, V.R.: Efficient
  {GAN}-based anomaly detection. arXiv preprint arXiv:1802.06222  (2018)

\bibitem{zhou2016learning}
Zhou, B., Khosla, A., Lapedriza, A., Oliva, A., Torralba, A.: Learning deep
  features for discriminative localization. In: Proceedings of the IEEE
  conference on computer vision and pattern recognition. pp. 2921--2929 (2016)

\end{thebibliography}
\end{document}